\documentclass{article}

\usepackage[preprint]{neurips_2026}

\makeatletter

\newif\ifpreprintmode
\newif\ifshowpubliccodeurl
\if@preprint
  \preprintmodetrue
  \showpubliccodeurltrue
\else
  \preprintmodefalse
  \if@neuripsfinal
    \showpubliccodeurltrue
  \else
    \showpubliccodeurlfalse
  \fi
\fi
\newcommand{\codeurl}{https://example.com/anonymous-repository}
\newcommand{\codefootnote}{\footnote{Code is available at \url{\codeurl}.}}

\ifshowpubliccodeurl
  \renewcommand{\codeurl}{https://github.com/cuent/scfm}
  
\fi
\makeatother

\usepackage[utf8]{inputenc} 
\usepackage[T1]{fontenc}    
\usepackage{hyperref}       
\usepackage{url}            
\usepackage{booktabs}       
\usepackage{amsfonts}       
\usepackage{nicefrac}       
\usepackage{microtype}      
\usepackage{xcolor}         
\usepackage{import}
\usepackage{amsmath}
\usepackage{amsthm}
\usepackage{wrapfig}
\usepackage{graphicx}
\usepackage{subcaption}
\usepackage{todonotes}
\usepackage{etoc}
\usepackage{algorithm}
\usepackage{algpseudocode}
\usepackage{multirow}
\usepackage{enumitem}
\usepackage{makecell}
\newtheorem{proposition}{Proposition}[section]

\title{Structured Coupling for Flow Matching}

\author{%
  Xavier Sumba, Carles Balsells-Rodas, Yingzhen Li \\
  Imperial College London\\
  \texttt{\{xxs22,cb221,yingzhen.li\}@imperial.ac.uk} \\
}

\begin{document}

\maketitle

\begin{abstract}
  Standard flow matching scales well but typically relies on an unstructured source distribution, limiting its ability to learn interpretable latent structure.
Latent-variable models, by contrast, capture structure but often sacrifice generative quality.
We bridge this gap by proposing Structured Coupling for Flow Matching (SCFM), a cooperative framework that augments flow matching with structured latent representation learning. By introducing structured latent variables and exogenous noise into the source,
SCFM jointly learns a structured prior (via latent variable modeling) and a continuous transport map (via flow matching). It uses a shared time-dependent recognition network for both latent variable model variational inference and intermediate-time flow velocity estimation.
This yields a structurally informed yet unconditional, simulation-free flow model, where the latent variable model can also assist flow sampling.
Empirically, SCFM facilitates unsupervised latent representation learning for clustering, disentanglement and downstream tasks, while remaining competitive with flow matching in sample quality, showing that meaningful structure can be learned without sacrificing generative fidelity\codefootnote.

\end{abstract}

\etocdepthtag.toc{mtchapter}

\section{Introduction}
Diffusion models \citep{ho2020denoising,nichol2021improved,songdenoising,songscore} and flow-based models \citep{lipman2023flow,liu2022flow,gat2024discrete,tong2024improving,isobe2025extended} are central paradigms in modern generative modeling.
Among flow-based approaches, flow matching has emerged as an effective framework for training continuous normalizing flows \citep{lipman2023flow,NEURIPS2024_15b78035,albergo2025stochastic}, combining scalable optimization with strong sample quality \citep{ma2024sit}.
However, standard flow matching typically uses a fixed, unstructured source distribution, so the learned transport does not explicitly encourage latent structure useful for interpretation, clustering, disentanglement, or downstream tasks.

Latent-variable models such as variational autoencoders (VAEs) \citep{Kingma2014,burda2015importance} address the complementary problem.
By introducing an explicit latent space and amortized inference, they can learn structured and often interpretable representations, including representations that better separate underlying factors of variation \citep{dilokthanakul2016deep,10.5555/3172077.3172161,higgins2017betavae,chen2018isolating,pmlrv97locatello19a}.
However, their generative quality is often limited by restrictive decoder families or posterior approximations \citep{burgess2018understanding}.
This leaves a gap between methods that learn useful latent structure and methods that produce high-fidelity samples in a single, unified framework.

\begin{table*}[t]
    \centering
    \scriptsize
    \setlength{\tabcolsep}{3pt}
    \renewcommand{\arraystretch}{1.05}
    \caption{
        Standard flow matching (FM) versus structured coupling for flow matching (SCFM).
    }
    \label{tab:intro_comparison}
    \begin{tabular*}{\textwidth}{@{\extracolsep{\fill}}p{0.16\textwidth}p{0.22\textwidth}p{0.55\textwidth}@{}}
        \toprule
        \textbf{Aspect} & \textbf{FM}                                                                               & \textbf{SCFM} \\
        \midrule
        Source          & fixed $p_0(\mathbf{x}_0)$
        & $\mathbf{x}_0=(\mathbf{z},\varepsilon)$, $p_\psi(\mathbf{z})p(\varepsilon)$                               \\
        Train coupling
        & $p_0(\mathbf{x}_0)\,p_{\mathrm{data}}(\mathbf{x}_1)$
        & $p_{\mathrm{data}}(\mathbf{x}_1)\,q_\phi(\mathbf{z}\!\mid\!\mathbf{x}_1)\,p(\varepsilon)$ ($\approx p_{\theta}(\mathbf{x}_1 \!\mid\!\mathbf{z})p_{\psi}(\mathbf{z})p(\varepsilon)$)                 \\
        Interpolant     & $I_t(\mathbf{x}_0,\mathbf{x}_1)$
        & same linear $I_t$                                                                                         \\
        Posterior model & none explicit
        & shared $q_{t,\phi}(\mathbf{x}_0\mid\mathbf{x}_t)$; at $t=1$, recovers VAE posterior $q_\phi(\mathbf{z}\!\mid\!\mathbf{x}_1)$       \\
        Sampling       & flow from fixed $p_0(\mathbf{x}_0)$
        & flow from $p_\psi(\mathbf{z})p(\varepsilon)$; optional decoder proposal plus short refinement              \\
        Endpoint branch & none
        & VAE-style prior and decoder learning                                                                      \\
        Latent structure       & none explicit
        & structured latents for clustering, disentanglement, and downstream tasks                                  \\
        \bottomrule
    \end{tabular*}
\end{table*}

\begin{figure*}[t]
    \centering
    \includegraphics[width=\textwidth]{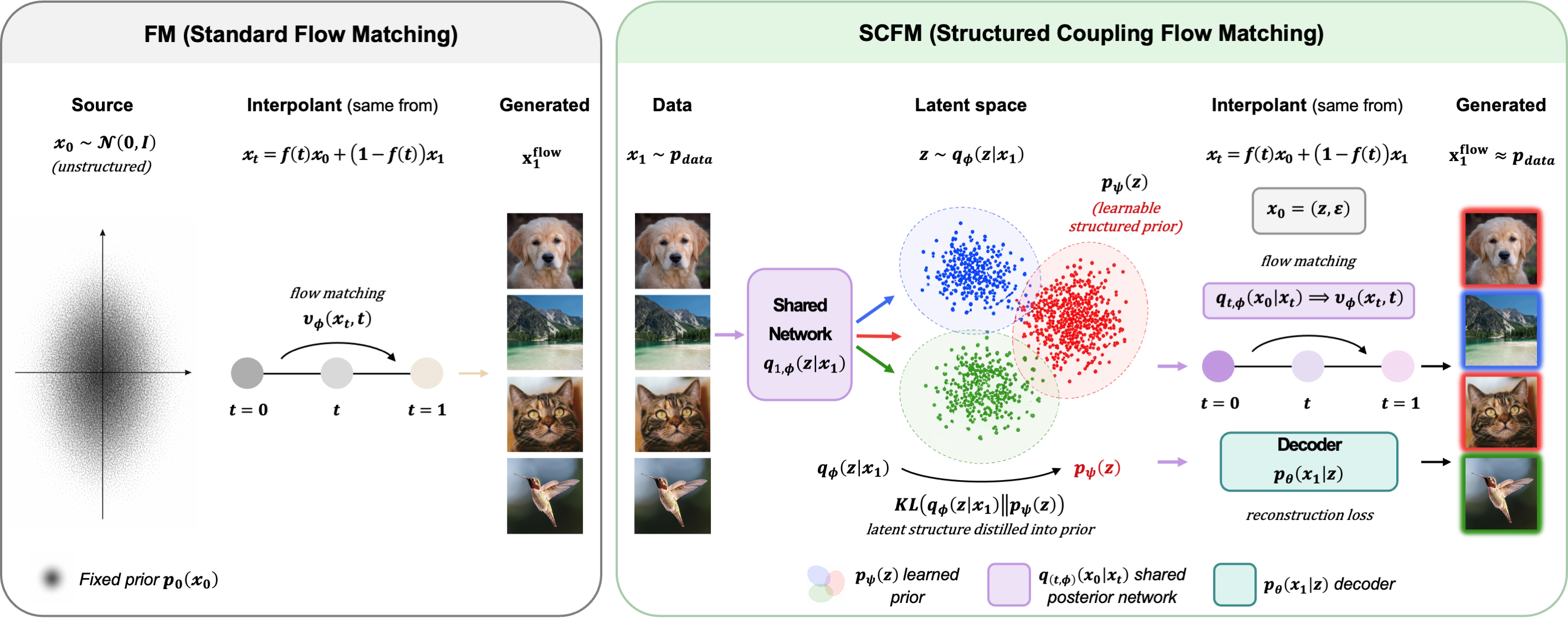}
    \caption{
        While standard flow matching (left) relies on a fixed, unstructured source, SCFM (right) replaces this with an augmented source $\mathbf{x}_0=(\mathbf{z},\varepsilon)$. An encoder induces a data-dependent coupling, distilling semantic structure into a learnable prior (indicated by colored latent clusters).
        Crucially, a single shared network serves as both the endpoint variational encoder for prior learning and the intermediate-time posterior estimator for flow velocity.
        This unified objective yields a structured latent space without sacrificing the continuous, simulation-free transport of standard flow models.
    }
    \label{fig:intro_comparison}
\end{figure*}

To close this gap, we propose \emph{Structured Coupling for Flow Matching} (SCFM), a framework that combines structured latent-variable learning with simulation-free flow matching in a stochastic-interpolant formulation \citep{albergo2023building,albergo2025stochastic}.
SCFM replaces the standard source with an augmented variable $\mathbf{x}_0=(\mathbf{z},\varepsilon)$, where $\mathbf{z}$ follows a learnable structured prior and $\varepsilon$ provides exogenous transport degrees of freedom.
During training, an encoder induces the coupling between data and source variables, while a shared time-dependent recognition network acts as the variational encoder at the endpoint ($t=1$) and as the posterior mean estimator that defines the flow for intermediate times ($t<1$).
A VAE-style endpoint objective aligns the aggregated posterior over $\mathbf{z}$ with the prior, so the flow is trained and sampled from the same structured source distribution.
Through this VAE objective, the flow becomes structurally informed while remaining unconditional, enabling reconstruction, unconditional generation, and decoder-initialized refinement. SCFM therefore combines the representation-learning benefits of latent-variable models with the sample quality of flow matching.

Our contributions are summarized as follows (also see Figure~\ref{fig:intro_comparison} and Table~\ref{tab:intro_comparison}):
\begin{itemize}[itemsep=0em]
    \vspace{-0.5em}
    \item We introduce SCFM, a structured flow-matching framework that jointly learns a latent prior and a continuous transport map. The core idea is to use an augmented source variable $\mathbf{x}_0=(\mathbf{z},\varepsilon)$, together with an encoder-induced coupling that separates semantic structure from transport flexibility.
    \item We show how a shared time-dependent recognition network and VAE-style endpoint objective align the training coupling with the sampling prior, while also enabling reconstruction and decoder-initialized refinement from endpoint proposals. Our approach integrates flow matching and VAE training in a cooperative framework, avoiding two-stage or independent training of different network components.
    \item We demonstrate empirically that SCFM learns structural latent representations that are useful for clustering, disentanglement, and downstream classification, while preserving competitive sample generation quality. The decoder-initialized refinement method also improves VAE generation quality towards the levels of flow-matching models.
\end{itemize}



\section{Preliminaries}
\label{sec:preliminaries}

\paragraph{(Variational) Flow matching.} Continuous normalizing flows (CNFs) \citep{chen2018neural} transforms samples from a source distribution into data by solving an ordinary differential equation:
\begin{equation}
    \frac{d\mathbf{x}_t}{dt}=v_{\phi, t}(\mathbf{x}_t),
    \qquad
    \mathbf{x}_0\sim p_0(\mathbf{x}_0),\quad t\in[0,1].
    \label{eq:cnf_ode}
\end{equation}
Likelihood-based CNF training requires tracking the change of density along the probability path $\{p_t\}_{t\in[0,1]}$, which introduces numerical integration and Jacobian trace terms \citep{chen2018neural,grathwohl2018scalable}.
Flow matching \citep{lipman2023flow,albergo2023building,albergo2025stochastic,ma2024sit} avoids these costs by learning $v_{\theta,t}$ through regression against velocities induced by a prescribed path between source and target distributions.
Once the vector field is learned, sampling simply draws $\mathbf{x}_0\sim p_0$ and integrates the ODE \eqref{eq:cnf_ode} forward to $t=1$, producing samples that are approximately distributed as the data distribution $p_{\text{data}}(\mathbf{x}_1)$.


In detail, let $\Gamma(\mathbf{x}_0,\mathbf{x}_1)$ be a coupling between a source distribution $p_0(\mathbf{x}_0)$ and the data distribution $p_{\text{data}}(\mathbf{x}_1)$.
With $f(t)=1-t$, the interpolant
\begin{equation}
    \mathbf{x}_t
    =
    f(t)\mathbf{x}_0 + (1-f(t))\mathbf{x}_1,
    \qquad t\in[0,1],
    \label{eq:interpolant}
\end{equation}
defines a path from source samples to data samples.
The corresponding conditional velocity field is
$v_t(\mathbf{x}_t\mid \mathbf{x}_0)
    =
    \frac{\partial_t f(t)}{1-f(t)}
    \bigl(\mathbf{x}_0-\mathbf{x}_t\bigr).$
The marginal velocity field that transports the marginal distribution of $\mathbf{x}_t$ is obtained by averaging this conditional velocity over the posterior source distribution $\Gamma_t(\mathbf{x}_0\mid\mathbf{x}_t)$, which is defined by the interpolant \eqref{eq:interpolant} and the coupling $\Gamma(\mathbf{x}_0, \mathbf{x}_1)$:
\begin{equation}
    v_t(\mathbf{x}_t)
    =
    \mathbb{E}_{\Gamma_t(\mathbf{x}_0\mid\mathbf{x}_t)}
    \!\left[
        v_t(\mathbf{x}_t\mid\mathbf{x}_0)
        \right]
    =
    \frac{\partial_t f(t)}{1-f(t)}
    \left(
    \mathbb{E}_{\Gamma_t(\mathbf{x}_0\mid\mathbf{x}_t)}
    [\mathbf{x}_0]
    -
    \mathbf{x}_t
    \right).
    \label{eq:marginal_velocity}
\end{equation}

Flow matching trains a neural network-based velocity field to approximate the marginal velocity field \eqref{eq:marginal_velocity} via regression. Instead, Variational Flow Matching (VFM) \citep{NEURIPS2024_15b78035} introduces a recognition model $q_{t,\phi}(\mathbf{x}_0\mid\mathbf{x}_t)$ to approximate $\Gamma_t(\mathbf{x}_0\mid\mathbf{x}_t)$,  
so the induced vector field depends only on the approximate posterior mean (see Appendix~\ref{app:vfm-view} for details):
\[
    v_{\phi,t}(\mathbf{x}_t)
    =
    \frac{\partial_t f(t)}{1-f(t)}
    \bigl(
    \mathbb{E}_{q_{t,\phi}(\mathbf{x}_0\mid\mathbf{x}_t)}
    [\mathbf{x}_0]
    -
    \mathbf{x}_t
    \bigr).
\]

\paragraph{Role of the source.}
In standard (variational) flow matching, the source distribution $p_0(\mathbf{x}_0)$ is usually fixed to a simple prior, such as an isotropic Gaussian.
This choice is convenient for sampling, but the source coordinates are not trained to expose semantic structure; representation structure, if present, is only implicit in the learned transport. As we shall see,
SCFM instead makes the source distribution learnable and structured by tying the source posterior in Eq.~\eqref{eq:marginal_velocity} to VAE-style endpoint losses at the data endpoint, while retaining the simulation-free training advantages of flow matching.

\paragraph{Variational autoencoders.}
VAEs \citep{Kingma2014} build latent variable models for generative modelling,  where such a model consists of a prior $p_{\psi}(\mathbf{z})$, potentially with learnable parameters $\psi$ for structural representation learning, and a stochastic decoder $p_\theta(\mathbf{x}_1\mid\mathbf{z})$.
Training is done via minimizing the negative Evidence Lower Bound (ELBO), computed using an approximate posterior (stochastic encoder) $q_\phi(\mathbf{z}\mid\mathbf{x}_1)$:
\begin{equation}
    \mathcal{L}_{\mathrm{VAE}}(\theta, \phi, \psi)
    =
    \mathbb{E}_{p_{\text{data}}(\mathbf{x}_1)}
    \Big[
    -\mathbb{E}_{q_\phi(\mathbf{z}\mid\mathbf{x}_1)}
    [\log p_\theta(\mathbf{x}_1\mid\mathbf{z})]
    +
    \mathrm{KL}\bigl(q_\phi(\mathbf{z}\mid\mathbf{x}_1) \,\|\, p_{\psi}(\mathbf{z})\bigr)
    \Big].
    \label{eq:vae_preliminaries}
\end{equation}
While VAEs are powerful for representation learning, they often yield lower sample quality than diffusion/flow-based models.
As we shall see, SCFM exploits the VAE framework to structurally inform the flow's source distribution, unifying the representation learning of VAEs with the high-fidelity generation of flow matching.

\section{Structured Coupling for Flow Matching}
\label{sec:method-scfm}

We introduce SCFM, whose key ideas include (1) the encoder-induced coupling (Section~\ref{sec:method-augmented-source}), (2) time-split posterior matching and practical training loss (Section~\ref{sec:method-variational}), and (3) flexible sampling modes (Section~\ref{sec:method-sampling}).
Table~\ref{tab:intro_comparison} provides a summary of SCFM innovations over standard FM. In essence, SCFM makes the source marginal structural and learnable via encoder-induced coupling, while retaining the same linear interpolant and simulation-free training target.



\subsection{Encoder- and Decoder-Induced Couplings}
\label{sec:method-augmented-source}

SCFM introduces structure by changing the source endpoint used by flow matching in order to achieve structural latent representation learning. We define the source variable as a concatenation of structural representation $\mathbf{z}$ and exogenous noise $\varepsilon$:
\[
    \mathbf{x}_0=(\mathbf{z},\varepsilon)\in\mathbb{R}^{D},
    \qquad
    \mathbf{z}\in\mathbb{R}^{d_z},
    \qquad
    \varepsilon\in\mathbb{R}^{d_\varepsilon},
    \qquad
    D=d_z+d_\varepsilon .
\]
We train a flow-matching generative model that, at sampling time, uses a learnable source prior factorized into a structural prior over $\mathbf{z}$ and a fixed exogenous-noise prior over $\varepsilon$:
\begin{equation}
    p_\psi(\mathbf{x}_0)
    =
    p_\psi(\mathbf{z})\,p(\varepsilon),
    \qquad
    p(\varepsilon)=\mathcal{N}(0,I_{d_\varepsilon}),
    \label{eq:structured_prior}
\end{equation}
where $p_\psi(\mathbf{z})$ is learnable.
In our experiments, \(p_\psi(\mathbf{z})\) is chosen to impose structure on the latent space through a Gaussian mixture prior.
Standard flow matching would typically use the independent coupling $p_\psi(\mathbf{x}_0)p_{\text{data}}(\mathbf{x}_1)$.
Instead, SCFM uses the \emph{encoder-induced coupling}
\begin{equation}
    \Gamma^{\text{enc}}_\phi(\mathbf{x}_0,\mathbf{x}_1)
    :=
    p_{\text{data}}(\mathbf{x}_1)\,
    q_\phi(\mathbf{z}\mid\mathbf{x}_1)\,
    p(\varepsilon),
    \qquad
    \mathbf{x}_0=(\mathbf{z},\varepsilon),
    \label{eq:structured_coupling}
\end{equation}
where $q_\phi(\mathbf{z}\mid\mathbf{x}_1)$ is a learnable stochastic encoder.
Thus, each interpolation path starts from a source endpoint whose structured coordinate is inferred
from the data endpoint. This makes the latent representation part of the transport problem itself,
rather than an auxiliary representation learned beside the flow.
Under this coupling, the optimal flow-matching model transports the source marginal
\begin{equation}
    \Gamma^{\text{enc}}_\phi(\mathbf{x}_0)
    =
    q_\phi^{\mathrm{agg}}(\mathbf{z})\,p(\varepsilon),
    \qquad
    q_\phi^{\mathrm{agg}}(\mathbf{z})
    :=
    \int
    p_{\text{data}}(\mathbf{x}_1)\,
    q_\phi(\mathbf{z}\mid\mathbf{x}_1)\,d\mathbf{x}_1,
    \label{eq:aggregated_posterior}
\end{equation}
to the data distribution $p_{\text{data}}(\mathbf{x}_1)$.
The resulting gap between the training marginal $\Gamma^{\text{enc}}_\phi(\mathbf{x}_0)$ and the sampling prior $p_\psi(\mathbf{x}_0)$ lies entirely in the structural prior over $\mathbf{z}$.
SCFM closes this prior-aggregated-posterior mismatch with VAE-style endpoint objective.
With a stochastic decoder $p_\theta(\mathbf{x}_1\mid\mathbf{z})$,  the VAE loss \eqref{eq:vae_preliminaries} is equivalent to
\begin{equation}
    \begin{aligned}
        \mathcal{L}_{\mathrm{VAE}}(\theta, \phi, \psi)
        = &
        \ \mathrm{KL}\bigl(
        p_{\text{data}}(\mathbf{x}_1)\,q_\phi(\mathbf{z}\mid\mathbf{x}_1)
        \,\|\,
        p_\psi(\mathbf{z})\,p_\theta(\mathbf{x}_1\mid\mathbf{z})
        \bigr)
        + \mathrm{const.}
    \end{aligned}
    \label{eq:vae_loss}
\end{equation}
At the global optimum of this objective, under sufficiently expressive networks and prior families \citep{hoffman2016elbo,pmlr-v80-alemi18a}, the aggregated posterior matches the learnable prior, $q_\phi^{\mathrm{agg}}(\mathbf{z})=p_\psi(\mathbf{z})$ (see Appendix~\ref{app:scfm-endpoint-consistency}).
Consequently, the coupling used to train the flow is aligned with the source prior used at sampling time.

Jointly training the flow and the latent-variable model has two complementary benefits:
\begin{enumerate}[leftmargin=1.5em,itemsep=0pt]
    \item \textbf{Flow matching enhances the latent-variable model through deterministic transport-based sampling.}
          When the KL in Eq.~\eqref{eq:vae_loss} is zero (i.e., the VAE loss is minimized to its global optimum), the encoder-induced coupling coincides with the following \emph{decoder-induced coupling}
          \[
              \Gamma^{\text{dec}}_{\theta,\psi}(\mathbf{x}_0,\mathbf{x}_1)
              :=
              p_\psi(\mathbf{z})\,p(\varepsilon)\,p_\theta(\mathbf{x}_1\mid\mathbf{z}),
              \qquad
              \mathbf{x}_0=(\mathbf{z},\varepsilon).
          \]
          Its $t=1$ marginal is the latent-variable model
          \[
              \Gamma^{\text{dec}}_{\theta,\psi}(\mathbf{x}_1)
              =
              p_{\theta,\psi}(\mathbf{x}_1)
              :=
              \int p_\psi(\mathbf{z})\,p_\theta(\mathbf{x}_1\mid\mathbf{z})\,d\mathbf{z}.
          \]
          Therefore SCFM supports an alternative deterministic sampling mode for generation:
          \begin{equation}
              \mathbf{x}_1 \sim p_{\theta,\psi}(\mathbf{x}_1)
              \quad \Leftrightarrow \quad
              \mathbf{x}_0=(\mathbf{z},\varepsilon)\sim p_\psi(\mathbf{z})\,p(\varepsilon),
              \qquad
              \mathbf{x}_1=\mathbf{x}_0+\int_0^1 v_t(\mathbf{x}_t)\,dt.
          \end{equation}
          This connection motivates the decoder-initialized refinement scheme in Section~\ref{sec:method-sampling}.

    \item \textbf{VAE enriches flow-matching models via structured latent representation learning.}
          Beyond matching the prior to the aggregated posterior, the encoder $q_\phi(\mathbf{z}\mid\mathbf{x}_1)$ is trained to approximate the posterior of the latent-variable model $p_{\theta,\psi}(\mathbf{x}_1,\mathbf{z})$, thereby extracting compressed representations of the data into $\mathbf{z}$.
          Through this VAE objective, the flow itself becomes structurally informed: both the learned source prior $p_\psi(\mathbf{x}_0)$ and the encoder-induced coupling $\Gamma^{\text{enc}}_\phi(\mathbf{x}_0,\mathbf{x}_1)$ that define flow training inherit the latent structure learned in $\mathbf{z}$.
          Importantly, this is not external conditioning. The flow remains unconditional, while $\mathbf{z}$ still acquires semantic meaning in an unsupervised manner, which later supports disentanglement, clustering, and latent-space classification, and provides a degree of control through the learned latent variable.
\end{enumerate}

\subsection{Time-split Posterior Matching and Training Objectives}
\label{sec:method-variational}
\label{sec:method-objective}

The remaining question is how to couple the endpoint latent-variable objective to flow matching without a second encoder.
In standard FM, under the linear interpolant Eq.~\eqref{eq:interpolant}, the marginal vector field depends on the posterior only through
\(\mathbb{E}_{\Gamma_t(\mathbf{x}_0\mid\mathbf{x}_t)}[\mathbf{x}_0]\).
Similarly, SCFM uses a single time-dependent posterior-mean network over the structured source endpoint \(\mathbf{x}_0=(\mathbf{z},\varepsilon)\).

\begin{proposition}[Structured source posterior velocity]
    \label{prop:structured-source-velocity}
    Under the encoder-induced coupling in Eq.~\eqref{eq:structured_coupling} and the linear interpolant in Eq.~\eqref{eq:interpolant}, the marginal velocity is
    \begin{equation}
        v_t(\mathbf{x}_t)
        =
        \frac{\partial_t f(t)}{1-f(t)}
        \left(
        \mathbb{E}_{\Gamma^{\text{enc}}_t(\mathbf{x}_0\mid\mathbf{x}_t)}[\mathbf{x}_0]
        -
        \mathbf{x}_t
        \right)
        =
        \frac{\partial_t f(t)}{1-f(t)}
        \left(
        \mathbb{E}_{\Gamma^{\text{enc}}_t(\mathbf{z},\varepsilon\mid\mathbf{x}_t)}
        [(\mathbf{z},\varepsilon)]
        -
        \mathbf{x}_t
        \right).
        \label{eq:structured_posterior_velocity}
    \end{equation}
\end{proposition}

\paragraph{Intermediate-time regime (\(t<1\)).}
For \(t<1\), SCFM reduces to variational flow matching on the structured source endpoint, using the fixed-covariance Gaussian approximate posterior (encoder)
\begin{equation}
    q_{t,\phi}(\mathbf{x}_0\mid\mathbf{x}_t)
    =
    \mathcal{N}
    \left(
    \mu_\phi(\mathbf{x}_t,t),
    \sigma_{\mathbf{x}_0}^{2}I_D
    \right) \quad \Rightarrow \quad v_{\phi,t}(\mathbf{x}_t)
    =
    \frac{\partial_t f(t)}{1-f(t)}
    \left(
    \mu_\phi(\mathbf{x}_t,t)-\mathbf{x}_t
    \right).
    \label{eq:unified_recognition}
\end{equation}
The induced vector field $v_{\phi,t}(\mathbf{x}_t)$ follows Eq.~\eqref{eq:structured_posterior_velocity} but with the oracle posterior mean replaced by the encoder mean.
The resulting posterior-matching objective is
\begin{equation}
    \mathcal{J}_{<1}(\phi)
    =
    \mathbb{E}_{t\sim\rho_{<1}}
    \mathbb{E}_{p_t(\mathbf{x}_t)}
    \mathrm{KL}
    \left(
    \Gamma^{\text{enc}}_t(\mathbf{x}_0\mid\mathbf{x}_t)
    \,\|\,
    q_{t,\phi}(\mathbf{x}_0\mid\mathbf{x}_t)
    \right),
    \label{eq:scfm_intermediate_kl}
\end{equation}
where \(\rho_{<1}\) is a distribution over $t$ supported on \(t<1\), and \(\Gamma^{\text{enc}}_t\) is induced by the encoder coupling in Eq.~\eqref{eq:structured_coupling}. As this matching objective is for training $q_{t, \phi}(\mathbf{x}_0 \mid \mathbf{x}_1)$ only, we apply stop-gradient operation $\text{sg}(\cdot)$ to $\Gamma^{\text{enc}}_t(\mathbf{x}_0 \mid \mathbf{x}_1)$ and use an equivalent compact VFM objective instead:
\begin{equation}
    \mathcal{L}_{\mathrm{VFM}}(\phi)
    =
    -\,\mathbb{E}_{t\sim\rho_{<1},(\mathbf{x}_t,\mathbf{x}_0)\sim \Gamma^{\text{enc}}_t(\mathbf{x}_0, \mathbf{x}_t)}
    \big[
        \log q_{t,\phi}(\text{sg}(\mathbf{x}_0) \mid \text{sg}(\mathbf{x}_t))
        \big]
    + \mathrm{const.},
    \label{eq:vfm_regression}
\end{equation}
where \((\mathbf{x}_t,\mathbf{x}_0)\) are sampled from the encoder interpolant-induced joint.
In practice, we draw \(\mathbf{x}_1\sim p_{\mathrm{data}}\), then \(\mathbf{z}\sim q_\phi(\mathbf{z}\mid\mathbf{x}_1)\) and \(\varepsilon\sim p(\varepsilon)\), set \(\mathbf{x}_0=(\operatorname{sg}(\mathbf{z}),\varepsilon)\), and form \(\mathbf{x}_t=f(t)\mathbf{x}_0+(1-f(t))\mathbf{x}_1\).
Under the fixed-covariance Gaussian family in Eq.~\eqref{eq:unified_recognition}, this KL is equivalent to posterior-mean regression, and hence to the usual time-weighted velocity regression (see Appendix~\ref{app:scfm-intermediate-vfm}).

\paragraph{Endpoint regime (\(t=1\)).}
The exact posterior $\Gamma^{\text{enc}}_t(\mathbf{x}_0 \mid \mathbf{x}_t)$ of the encoder-induced coupling converges to $q_{\phi}(\mathbf{z} \mid \mathbf{x}_1) p(\varepsilon)$ as $t \rightarrow 1$, meaning that at \(t=1\) VFM training is no longer suitable. Instead, at \(t=1\) SCFM performs structural learning by minimising the VAE loss Eq.~\eqref{eq:vae_loss}, together with decoder training.
In particular, the convergence result inspires us to use the \emph{same} network for the mean $\mu_\phi(\mathbf{x}_t,t)$ for both $t < 1$ and $t=1$. Furthermore, via splitting $\mu_\phi(\mathbf{x}_1,1) = \left(\mu_\phi^z(\mathbf{x}_1),\mu_\phi^\varepsilon(\mathbf{x}_1)\right)$, the first \(d_z\) coordinates of the mean network, together with an endpoint-only variance head branching from the mean network, parameterize the approximate posterior in the VAE
\begin{equation}
    q_\phi(\mathbf{z}\mid\mathbf{x}_1)
    =
    \mathcal{N}
    \left(
    \mu_\phi^z(\mathbf{x}_1),
    \operatorname{diag}(\sigma_\phi^2(\mathbf{x}_1))
    \right).
\end{equation}
The exogenous variable $\varepsilon$ is excluded from the decoder $p_{\theta}(\mathbf{x}_1 \mid \mathbf{z})$. Again due to the exact posterior convergence result,
by defining \(q_\phi^\varepsilon(\varepsilon\mid\mathbf{x}_1)=\mathcal{N}(\mu_\phi^\varepsilon(\mathbf{x}_1),I_{d_\varepsilon})\), we minimize the KL divergence
\begin{equation}
    \mathcal{R}_{\varepsilon}(\phi)
    =
    \frac{1}{2}
    \mathbb{E}_{p_{\text{data}}(\mathbf{x}_1)}
    \left[ \left\|
        \mu_\phi^\varepsilon(\mathbf{x}_1)
        \right\|^2 \right] = \mathrm{KL}(q_\phi^\varepsilon(\varepsilon \mid\mathbf{x}_1)\,\|\,p(\varepsilon)) + \text{const} .
    \label{eq:exogenous_regularizer}
\end{equation}
Thus at $t = 1$ the endpoint contribution to the total loss is $\mathcal{L}_{\mathrm{end}} = \mathcal{L}_{\mathrm{VAE}} + \mathcal{R}_{\varepsilon}$.
%
Appendix~\ref{app:scfm-endpoint-objective} gives the endpoint-objective derivation.

\paragraph{Flow--encoder network sharing.}
The same mean network \(\mu_\phi(\mathbf{x}_t,t)\) serves as the intermediate-time posterior estimator and, at \(t=1\), as the encoder mean \(\mu_\phi^z(\mathbf{x}_1)\); only the encoder variance head is endpoint-specific.
The endpoint VAE term also encourages \(q_\phi^{\mathrm{agg}}(\mathbf{z})\approx p_\psi(\mathbf{z})\), making the encoder-induced training marginal compatible with the sampling prior; see Eq.~\eqref{eq:vae_loss} and Section~\ref{sec:method-augmented-source}. This is a tractable regularizer rather than a guarantee of exact marginal alignment.

\paragraph{Total loss objective.}
In practice, one SCFM training step, summarized in Algorithm~\ref{alg:scfm_training} in appendix, optimizes the decomposed objective
\begin{equation}
    \mathcal{L}_{\mathrm{SCFM}}(\theta,\phi,\psi)
    =
    \mathcal{L}_{\mathrm{VFM}}(\phi)
    +
    \mathcal{L}_{\mathrm{rec}}(\theta,\phi)
    +
    \mathcal{L}_{\mathrm{KL}}(\phi,\psi)
    +
    \mathcal{R}_{\varepsilon}(\phi).
    \label{eq:scfm_objective}
\end{equation}

Here \(\mathcal{L}_{\mathrm{VFM}}\) trains the shared posterior model away from the endpoint, \(\mathcal{L}_{\mathrm{rec}}+\mathcal{L}_{\mathrm{KL}} = \mathcal{L}_{\mathrm{VAE}}\) is the endpoint VAE term, and \(\mathcal{R}_{\varepsilon}\) anchors the exogenous coordinates.
We use $\beta$-VAE~~\citep{higgins2017betavae} or $\beta$-TCVAE~~\citep{chen2018isolating} endpoint losses to enforce structured latent learning under a GMM prior. Reconstruction is trained with either a perceptual loss or MSE; see Appendix~\ref{app:scfm-practical-training}.

\subsection{Sampling via Decoder-Initialized Refinement}
\label{sec:method-sampling}

SCFM supports two flow-based sampling modes. The standard mode (Algorithm~\ref{alg:scfm_sampling} in appendix) is full ODE integration, similar to standard flow matching. For unconditional generation, we draw \(\mathbf{z}\sim p_\psi(\mathbf{z})\) and \(\varepsilon\sim p(\varepsilon)\), set \(\mathbf{x}_0=(\mathbf{z},\varepsilon)\), and integrate $v_{\phi,t}(\mathbf{x}_t)$ from \(t=0\) to \(t=1\). As SCFM also trains an encoder, if an observation \(\mathbf{x}_1\) is available, the same procedure yields a reconstruction by replacing the prior draw with \(\mathbf{z}\sim q_\phi(\mathbf{z}\mid\mathbf{x}_1)\) while still sampling \(\varepsilon\sim p(\varepsilon)\).

The decoder in SCFM provides an alternative sampling mode, as it can be used to initialize the flow near the endpoint. This \emph{decoder-initialized refinement} mode is summarized in Algorithm~\ref{alg:scfm_fast_sampling} in appendix. Given \(\mathbf{x}_0=(\mathbf{z},\varepsilon)\), we first sample \(\widehat{\mathbf{x}}_1\sim p_\theta(\cdot\mid\mathbf{z})\) and then initialize the interpolant at some \(t_0\in[0,1)\) via $\mathbf{x}_{t_0} = f(t_0)\mathbf{x}_0 + (1-f(t_0))\widehat{\mathbf{x}}_1$.
We then integrate $v_{\phi,t}(\mathbf{x}_t)$ only on \([t_0,1]\), trading fewer ODE evaluations for a stronger dependence on decoder quality. This is motivated by the fact that at the global optimum of VAE training we have $\Gamma^{\text{dec}}_{\theta,\psi}(\mathbf{x}_0,\mathbf{x}_1) = \Gamma^{\text{enc}}_{\phi}(\mathbf{x}_0,\mathbf{x}_1)$, so that flow sampling techniques are also applicable to decoder-induced couplings.

\section{Experiments}
\label{sec:experiments}

\setlength{\textfloatsep}{8pt plus 1pt minus 2pt}
\setlength{\dbltextfloatsep}{8pt plus 1pt minus 2pt}
\setlength{\floatsep}{8pt plus 1pt minus 2pt}
\setlength{\dblfloatsep}{8pt plus 1pt minus 2pt}
\setlength{\intextsep}{8pt plus 1pt minus 2pt}
\setlength{\abovecaptionskip}{4pt}
\setlength{\belowcaptionskip}{0pt}

We evaluate SCFM along three axes: structured representations, disentanglement, and sample quality. We report quantitative metrics and qualitative visualizations; Appendix~\ref{app:metrics} defines all metrics.

\paragraph{Experimental setups.}

All SCFM models use a learnable GMM prior over $\mathbf{z}$. MNIST~\citep{lecun2002gradient} clustering and Cars3D~\citep{reed2015deep}/Shapes3D~\citep{kim2018disentangling} disentanglement use separate models with $\beta$-VAE and $\beta$-TCVAE endpoint losses; Appendices~\ref{app:mnist_setup} and~\ref{app:cars3d_shapes3d_setup} give the corresponding setups. In contrast, CIFAR-10~\citep{krizhevsky2009learning} and ImageNet-128~\citep{ILSVRC15} reuse the same trained SCFM model for both latent evaluation and image generation. CIFAR-10 uses a U-Net backbone with a $\beta$-VAE endpoint objective augmented by LPIPS~\citep{johnson2016perceptual,zhang2018unreasonable}; Appendix~\ref{app:cifar10_setup} gives the training details. ImageNet-128 follows a latent-diffusion-style setup: a pretrained Stable-Diffusion VAE maps images to VAE latents, and SCFM is trained directly in this latent image space using a SiT-XL/2 backbone~\citep{rombach2022high,ma2024sit}. Appendix~\ref{app:imagenet128_setup} gives the ImageNet-128 model, training, and sampling setup, and Appendix~\ref{app:cifar_imagenet_probe_setup} gives the probing setup.

\subsection{Structured Latent Representations}
\label{sec:exp-structured-latents}

\begin{figure*}[t]
    \centering
    \begin{minipage}[c]{0.33\textwidth}
        \centering
        \makebox[\linewidth][c]{%
            \includegraphics[width=\linewidth]{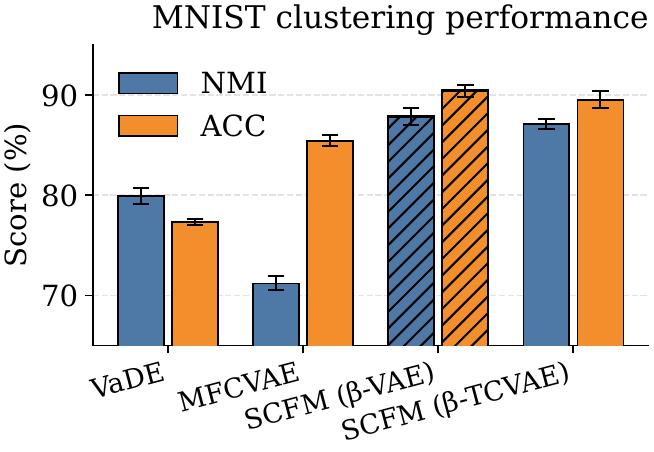}%
        }
    \end{minipage}\hfill
    \begin{minipage}[c]{0.33\textwidth}
        \centering
        \makebox[\linewidth][c]{%
            \includegraphics[width=\linewidth]{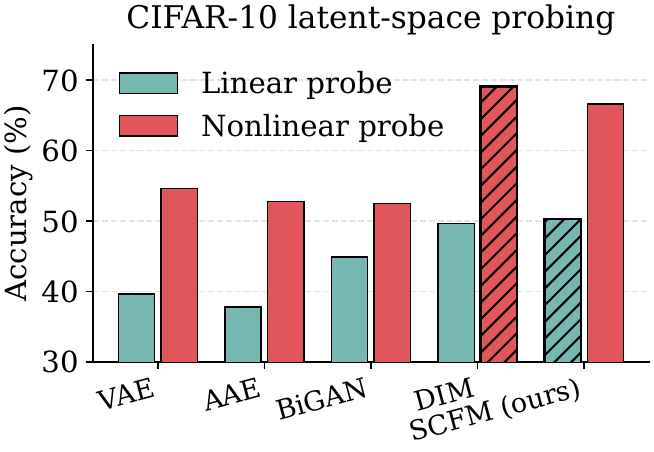}%
        }
    \end{minipage}\hfill
    \begin{minipage}[c]{0.33\textwidth}
        \centering
        \makebox[\linewidth][c]{%
            \includegraphics[width=\linewidth]{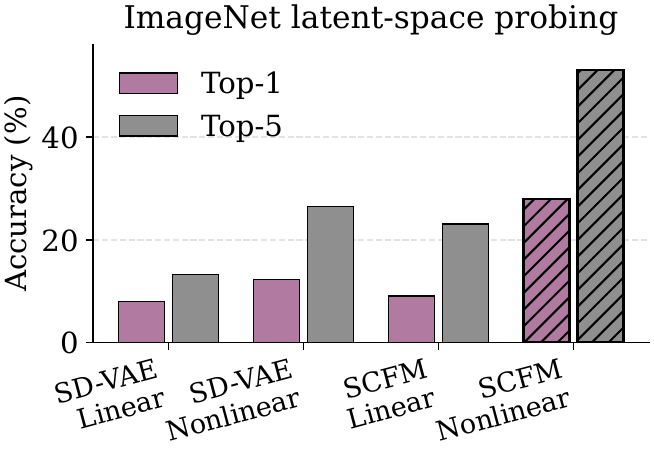}%
        }
    \end{minipage}

    \caption{\textbf{Structured latent representations.}
        Left: MNIST clustering metrics, reported as mean with standard-deviation error bars over five runs.
        Middle: CIFAR-10 downstream probe accuracy from learned latent representations.
        Right: ImageNet latent-space probing with frozen representations.
    }
    \label{fig:structured_latent_summary}
\end{figure*}

We first evaluate the representations learning capabilities of SCFM. On MNIST we measure cluster alignment, while on CIFAR-10 and ImageNet-128 we measure how much class information remains accessible from frozen latents without label supervision by training probing classifiers post-hoc.

\paragraph{MNIST clustering.}
On MNIST, we use a learnable GMM prior with $K=10$ and compare SCFM against VaDE~\citep{jiang2017variational} and MFCVAE~\citep{falck2021multi}. SCFM is trained with $\beta$-VAE~\citep{higgins2017betavae} and $\beta$-TCVAE~\citep{chen2018isolating} endpoint losses. We report normalized mutual information (NMI) and clustering accuracy (ACC) over five runs.

Figure~\ref{fig:structured_latent_summary} (left) summarizes the results, with full values in Appendix Table~\ref{tab:mnist_results}. SCFM ($\beta$-VAE) performs best, followed by SCFM ($\beta$-TCVAE). Relative to VaDE, the best SCFM variant improves NMI by nearly $8$ points and ACC by more than $13$ points.

\begin{figure*}[t!]
    \centering

    \begin{minipage}[c]{0.61\textwidth}
        \centering
        \includegraphics[width=\linewidth]{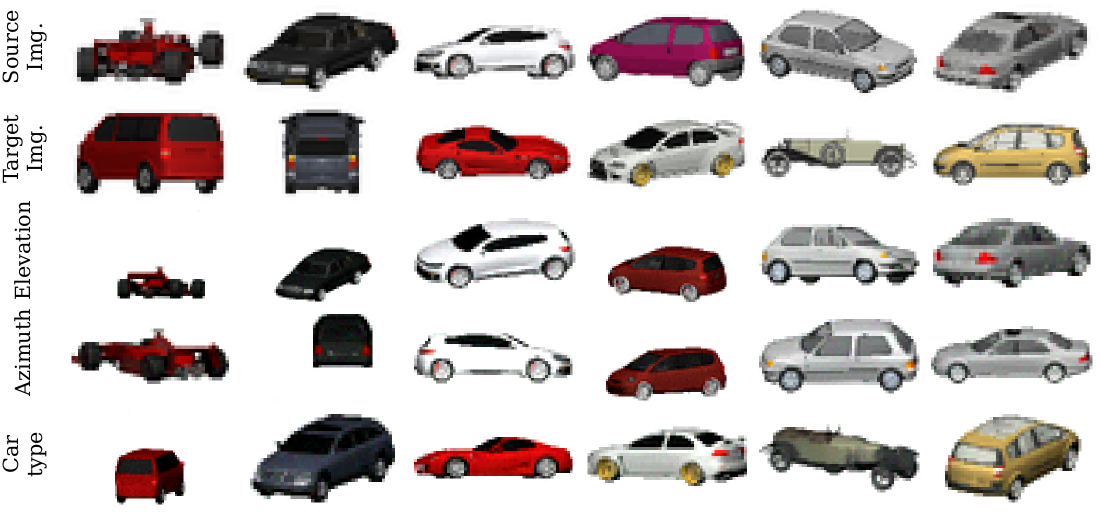}
    \end{minipage}\hfill
    \begin{minipage}[c]{0.36\textwidth}
        \centering
        \includegraphics[width=\linewidth]{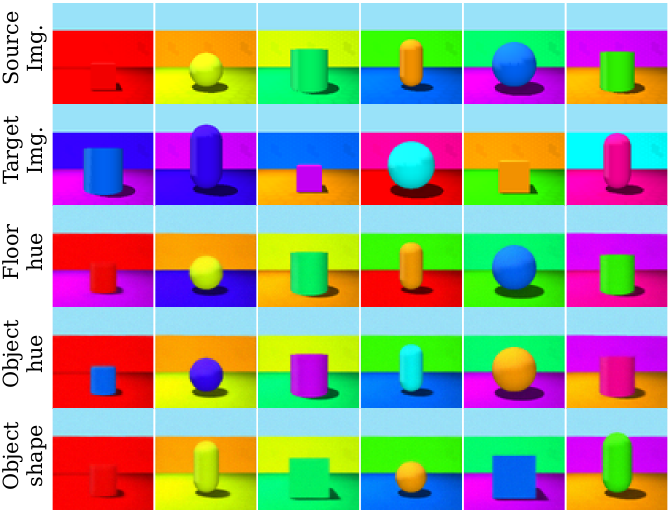}
    \end{minipage}

    \captionof{figure}{\textbf{Qualitative disentanglement via factor swaps.}
        Left: Cars3D factor swaps. Right: Shapes3D factor swaps. Each column shows one source-target pair; each swap row transfers one target factor while preserving the others.}
    \label{fig:disentangled_summary}

    \vspace{0.6em}

    \begin{minipage}{\textwidth}
        \centering
        \small
        \renewcommand{\arraystretch}{1.15}
        \setlength{\tabcolsep}{7pt}
        \begin{tabular}{l l cc cc}
            \toprule
             &                               & \multicolumn{2}{c}{\textbf{Cars3D}} & \multicolumn{2}{c}{\textbf{Shapes3D}}                                                              \\
            \cmidrule(lr){3-4} \cmidrule(lr){5-6}
             & \textbf{Method}               & \textbf{FactorVAE$\uparrow$}        & \textbf{DCI$\uparrow$}
             & \textbf{FactorVAE$\uparrow$}  & \textbf{DCI$\uparrow$}                                                                                                                   \\
            \midrule
            \multirow{2}{*}{\rotatebox[origin=c]{90}{VAE}}
             & $\beta$-VAE                   & 0.887 $\pm$ 0.039                   & 0.218 $\pm$ 0.045                     & 0.883 $\pm$ 0.091          & 0.624 $\pm$ 0.122             \\
             & $\beta$-TCVAE                 & 0.855 $\pm$ 0.082                   & 0.140 $\pm$ 0.019                     & 0.873 $\pm$ 0.074          & 0.613 $\pm$ 0.114             \\
            \midrule
            \multirow{4}{*}{\rotatebox[origin=c]{90}{Diffusion}}
             & DisDiff                       & \underline{0.976 $\pm$ 0.018}       & 0.232 $\pm$ 0.019                     & 0.902 $\pm$ 0.043          & 0.723 $\pm$ 0.013             \\
             & FDAE                          & 0.912 $\pm$ 0.020                   & 0.329 $\pm$ 0.061                     & 0.998 $\pm$ 0.003          & 0.762 $\pm$ 0.064             \\
             & EncDiff                       & 0.948 $\pm$ 0.017                   & \textbf{0.357 $\pm$ 0.072}            & \textbf{0.999 $\pm$ 0.001} & \textbf{0.952 $\pm$ 0.028}    \\
             & DyGA                          & 0.846 $\pm$ 0.015                   & 0.307 $\pm$ 0.032                     & 0.958 $\pm$ 0.044          & \underline{0.833 $\pm$ 0.054} \\
            \midrule
            \multirow{2}{*}{\rotatebox[origin=c]{90}{Flow}}
             & SCFM ($\beta$-VAE)            & 0.940 $\pm$ 0.019                   & \underline{0.337 $\pm$ 0.082}         & 0.836 $\pm$ 0.034          & 0.779 $\pm$ 0.065             \\
             & SCFM ($\beta$-TCVAE)          & \textbf{0.977 $\pm$ 0.027}          & \textbf{0.357 $\pm$ 0.078}
             & \underline{0.957 $\pm$ 0.083} & 0.828 $\pm$ 0.054                                                                                                                        \\
            \bottomrule
        \end{tabular}

        \captionof{table}{\textbf{Quantitative comparison on Cars3D and Shapes3D using FactorVAE score and DCI (mean$\pm$std).}
            For SCFM, we report both $\beta$-VAE and $\beta$-TCVAE endpoint regularizers. Best results are shown in \textbf{bold} and second-best results are \underline{underlined}.}
        \label{tab:disentangled_fm}
    \end{minipage}
\end{figure*}

\begin{wrapfigure}[13]{r}{0.33\linewidth}
    \vspace{-2em}
    \centering
    \includegraphics[width=\linewidth]{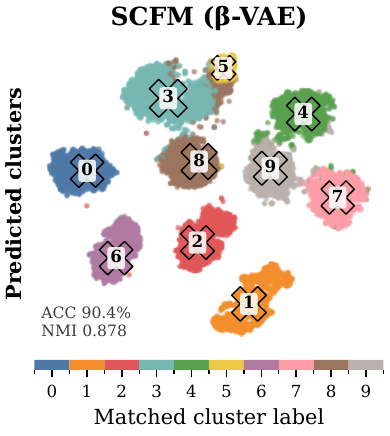}
    \caption{MNIST latents.}
    \label{fig:mnist_latent_main}
    \vspace{-0em}
\end{wrapfigure}
Figure~\ref{fig:mnist_latent_main} shows a largely label-aligned latent partition under the learned GMM, with most ambiguity between visually similar digits such as $3$ and $5$. Overall, $\mathbf{z}$ learns cluster-discriminative structure rather than acting only as a sampling coordinate.
Appendix~\ref{app:mnist_latent_diagnostics} provides the full comparison across models.

\vspace{-1em}
\paragraph{CIFAR-10 representation quality.}
On CIFAR-10, we set $K=10$ as a coarse structured prior, without assuming a one-to-one correspondence between mixture components and class labels. We therefore evaluate representation quality by freezing $\mathbf{z}$ and training linear and nonlinear probes without data augmentation. SCFM results are averaged over five seeds, while baseline numbers are taken from \citet{zhang2022improvingvaebasedrepresentationlearning}.

Figure~\ref{fig:structured_latent_summary} (middle) shows that SCFM achieves the best linear probe accuracy and the second-best nonlinear probe accuracy; full values are in Appendix Table~\ref{tab:cifar_compare}. It outperforms VAE~\citep{Kingma2014}, AAE~\citep{makhzani2015adversarial}, and BiGAN~\citep{donahue2016adversarial} on both probes while remaining competitive with DIM~\citep{hjelm2018learning}.
The source therefore retains class-relevant information despite being trained without labels.
Appendix~\ref{app:cifar10_latent_probe_diagnostics} shows reliable separation of vehicle classes, with most confusion among visually similar animal classes; Appendix~\ref{app:cifar10_generation} visualizes samples across mixture-components with coherent appearance statistics.


\paragraph{ImageNet-128 representation quality.}
We next test whether these representation gains persist at ImageNet scale. SCFM is trained in the Stable-Diffusion VAE latent space with a $K=100$ GMM prior, and we evaluate frozen $\mathbf{z}$ with linear and nonlinear probes.

Figure~\ref{fig:structured_latent_summary} (right) summarizes the ImageNet-128 results, with full values in Appendix Table~\ref{tab:imagenet_latent_probe_scfm}. Relative to a frozen SD-VAE~\citep{rombach2022high} encoder, SCFM improves linear Top-1/Top-5 accuracy from $8.00/13.23$ to $9.07/23.14$ and nonlinear Top-1/Top-5 accuracy from $12.34/26.54$ to $27.96/53.08$.
These gains indicate that the structured latent variable $\mathbf{z}$ retains substantially more class-accessible information than the pretrained VAE latent baseline.

\subsection{Disentanglement}
\label{sec:exp-disentanglement}

To test whether the learned structured latent variable supports controllable generative factors, we evaluate SCFM on Cars3D and Shapes3D, using ground-truth factors only for evaluation. Following \citet{pmlrv97locatello19a}, models are trained without factor labels, with latent dimension $10$ and a learnable GMM prior with $K=10$ components. For each endpoint regularizer, we train $10$ models while sweeping $\beta$; qualitative swaps are taken from the best run.

Table~\ref{tab:disentangled_fm} reports FactorVAE and DCI disentanglement scores, with VAE and diffusion baselines from \citet{chi2026disentangled}. SCFM is competitive on both datasets. On Cars3D, SCFM ($\beta$-TCVAE) attains the best FactorVAE score and a DCI score comparable to the strongest baselines, while SCFM ($\beta$-VAE) remains competitive on both metrics. On Shapes3D, prior diffusion models perform best, but both SCFM variants remain strong in the fully unsupervised setting.
%
%
Figure~\ref{fig:disentangled_summary} shows qualitative factor swaps obtained by interpolating in latent space and generated with full ODE sampling. SCFM transfers target factors while largely preserving the remaining attributes, although cylinder-to-cube transitions remain harder. Overall, SCFM learns factor-sensitive latent structure.

\subsection{Image Generation}
\label{sec:exp-image-generation}

\begin{table*}[t]
    \caption{\textbf{Sample quality on CIFAR-10 and ImageNet-128.}
        Left: FID 50K on CIFAR-10 and ImageNet-128; lower is better. For ImageNet-128, \emph{cond.} denotes a class-conditional SiT-XL/2 baseline trained with ImageNet labels, while \emph{uncond.} removes label conditioning. Right: representative SCFM samples, with ImageNet-128 samples on the top row and CIFAR-10 samples below.}
    \label{tab:fid_compare}
    \centering
    \small
    \begin{minipage}[c]{0.42\textwidth}
        \centering
        \footnotesize
        \setlength{\tabcolsep}{3.6pt}
        \renewcommand{\arraystretch}{1.10}
        \begin{tabular}{@{}llc@{}}
            \toprule
            \textbf{Dataset} & \textbf{Model}     & \textbf{FID 50k} $\downarrow$ \\
            \midrule
            \multirow{2}{*}{CIFAR-10}
                             & Flow Matching      & 2.137                         \\
                             & SCFM (ours)        & \textbf{2.117}                \\
            \midrule
            \multirow{3}{*}{ImageNet-128}
                             & SiT-XL/2 (cond.)   & 17.243                        \\
                             & SiT-XL/2 (uncond.) & 26.349                        \\
                             & SCFM (ours)        & \textbf{17.180}               \\
            \bottomrule
        \end{tabular}
    \end{minipage}\hfill
    \begin{minipage}[c]{0.58\textwidth}
        \centering
        \includegraphics[width=\linewidth]{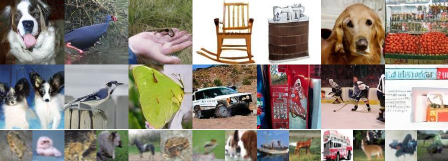}
    \end{minipage}
\end{table*}


\begin{wrapfigure}{r}{0.58\linewidth}
    \centering
    \vspace{-4.5em}
    \includegraphics[width=1\linewidth]{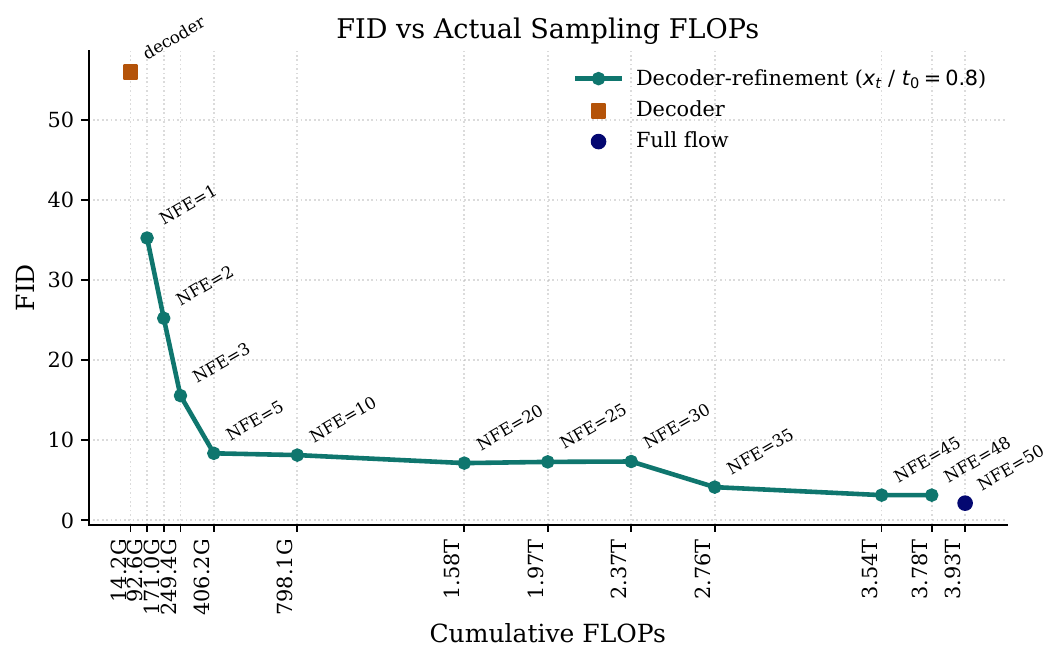}
    \caption{\textbf{CIFAR-10 FID vs. cumulative sampling FLOPs.}
        We compare decoder-only sampling, decoder refinement from $x_t$ with $t_0=0.8$, and full-flow sampling. Decoder refinement improves the quality--compute trade-off, approaching full-flow FID with fewer FLOPs.}
    \label{fig:cifar10_fid_vs_flops}
    \vspace{-1.em}
\end{wrapfigure}

Finally, we test whether adding a structured latent source preserves the sample quality of flow matching. Table~\ref{tab:fid_compare} reports FID 50K on CIFAR-10 and ImageNet-128 together with representative full-flow SCFM samples.

On CIFAR-10, SCFM reaches $2.117$ FID, essentially matching the $2.137$ flow-matching baseline~\citep{lipman2023flow}. Thus, learning structured source does not degrade sample quality.

We also evaluate decoder-initialized refinement for compute--quality trade-off. Figure~\ref{fig:cifar10_fid_vs_flops} shows that short refinement improves over decoder-only sampling with substantially fewer FLOPs; Appendix Figure~\ref{fig:cifar_component_rows} shows samples from all three modes.

On ImageNet-128, SCFM is unconditional at generation time, sampling $\mathbf{z}\sim p_\psi(\mathbf{z})$ rather than using class labels. It reaches $17.180$ FID, slightly improving over class-conditional SiT-XL/2 ($17.243$) and substantially outperforming unconditional SiT-XL/2 ($26.349$). This suggests that the learned structured source remains effective for large-scale unconditional generation, even at ImageNet scale.
Additional uncurated samples for all datasets are shown in Appendix~\ref{app:additional_qualitative}.

\section{Discussion}
\label{sec:discussion}
\textbf{Related work overview $\quad$}
Related work around SCFM falls into three main categories.
First, latent-variable models such as VAEs and their structured-prior extensions learn useful representations for clustering and disentanglement, but their generation quality is typically constrained by the decoder family and by direct sampling from the latent-variable model \citep{Kingma2014,burda2015importance,dilokthanakul2016deep,jiang2017variational,falck2021multi,higgins2017betavae,burgess2018understanding,chen2018isolating,pmlrv97locatello19a}.
Second, diffusion- and transport-based methods improve the source distribution or coupling geometry through learned priors, optimal-transport pairings, or variational couplings, but generally do not treat the source itself as a learned representation space \citep{lee2022priorgrad,10.5555/3618408.3618882,tong2024improving,pmlr-v235-albergo24a,2024arXiv241013431W,silvestri2025vct}.
Third, recent flow-matching approaches begin to target representation learning more directly through latent-variable transport, structured flow autoencoding, disentangled flows, or joint encoder-generator training \citep{guo2025variational,zhang2025towards,xu2026structured,ukita2026highperformance,chi2026disentangled}.
SCFM is closest to the last category, but differs in that it learns a structured \emph{source distribution} for a stochastic interpolant: the latent variable is part of the source endpoint being transported, not only a conditioning signal or a factorization of the velocity field.
Appendix~\ref{app:related-work} provides an extended related-work discussion.

\textbf{Comparison with guidance $\quad$}
%
SCFM is distinct from guidance-based generation. Classifier-free guidance and conditional flow-matching models steer sampling by injecting external conditioning signals, such as labels or prompts, into the denoising or velocity network~\citep{ho2022classifierfree}. SCFM instead samples unconditionally from $p_\psi(\mathbf{z})p(\varepsilon)$, without labels, prompts, or an external guidance term. The latent variable $\mathbf{z}$ can still shape generation because it is part of the source endpoint transported by the flow. Thus, SCFM provides implicit structural control through the learned source distribution, rather than explicit conditioning at sampling time.

\textbf{Summary and future work $\quad$}
SCFM combines a learnable structured latent prior with flow-based transport in a single generative framework. Across MNIST, CIFAR-10, Cars3D, Shapes3D, and ImageNet-128, it yields strong clustering and representation quality over VAE-based baselines while remaining competitive with strong flow-based generators in sample quality. These results support the claim that learning a structured source can make flow matching substantially more useful as a representation model without giving up its main generative strengths.
In particular, the same structured latent source that improves representation quality does not force a trade-off against generation quality.

\begin{table}[t]
    \centering
    \small
    \caption{\textbf{Model complexity.}
        We report trainable parameters and FLOPs for one batch-size-one forward pass through each model.}
    \label{tab:flow_flops}
    \begin{tabular}{lllcc}
        \toprule
        \textbf{Dataset} & \textbf{Backbone} & \textbf{Model} & \textbf{Params} & \textbf{FLOPs / forward} \\
        \midrule
        CIFAR-10         & U-Net             & Flow Matching  & 73.6M           & 40.4G                    \\
        CIFAR-10         & U-Net             & SCFM           & 101M            & 95.4G                    \\
        \midrule
        ImageNet-128     & SiT-XL/2          & Flow Matching  & 675M            & 58.1G                    \\
        ImageNet-128     & SiT-XL/2          & SCFM           & 1.1B            & 153G                     \\
        \bottomrule
    \end{tabular}
\end{table}


However, SCFM still has several limitations. First, it is more expensive than standard flow matching, since it introduces endpoint latent-variable components and an auxiliary decoder, increasing both parameter count and training FLOPs; Table~\ref{tab:flow_flops} quantifies this overhead which is notable but not unacceptable. Second, decoder-based sampling modes depend on the quality of the endpoint latent-variable model: if the latent model is poorly trained or suffers from posterior collapse, reconstruction quality degrades and decoder-initialized refinement becomes less effective. Future work should therefore focus on reducing this additional cost, improving decoder robustness, and extending SCFM to larger-scale, multimodal, and conditional regimes.



\bibliographystyle{plainnat}
\bibliography{bibliography}


\appendix
\newpage
\begin{center}
    \LARGE
    \textbf{Appendix for ``Structured Coupling for Flow Matching''}
\end{center}

\etocdepthtag.toc{mtappendix}
\etocsettagdepth{mtchapter}{none}
\etocsettagdepth{mtappendix}{subsection}
{\small \tableofcontents}
\clearpage
\paragraph{Appendix organization.}
Appendix~\ref{app:vfm-view} reviews the VFM background, and Appendix~\ref{app:scfm} gives the SCFM derivations and implementation details. Appendix~\ref{app:algorithms} summarizes training and sampling, Appendix~\ref{app:related-work} extends the related-work discussion, and Appendices~\ref{app:experimental_details}--\ref{app:additional_qualitative} collect experimental details, metrics, and additional results.

\section{Extended Variational Flow-Matching Background}
\label{app:vfm-view}

This appendix recalls the variational interpretation of flow matching introduced by \citet{NEURIPS2024_15b78035}, specialized to the source-endpoint convention used in the main text.
The key point is that the marginal vector field can be written as a posterior expectation over endpoints of the interpolant.
Approximating this posterior with a variational distribution yields a variational inference objective; under a Gaussian variational family and a linear interpolant, this objective reduces to the usual flow-matching regression loss.

Let $\Gamma(\mathbf{x}_0,\mathbf{x}_1)$ be a coupling between a source endpoint $\mathbf{x}_0$ and a data endpoint $\mathbf{x}_1$, and let
\[
    \mathbf{x}_t
    =
    f(t)\mathbf{x}_0+(1-f(t))\mathbf{x}_1 .
\]
For the source-to-data convention used in the main text, conditioning on $\mathbf{x}_0$ gives the conditional velocity
\begin{equation}
    v_t(\mathbf{x}_t\mid \mathbf{x}_0)
    =
    \frac{\partial_t f(t)}{1-f(t)}
    \bigl(\mathbf{x}_0-\mathbf{x}_t\bigr).
    \label{eq:app-vfm-cond-velocity}
\end{equation}
The marginal vector field is the posterior average of these conditional velocities,
\begin{equation}
    v_t(\mathbf{x}_t)
    =
    \mathbb{E}_{\Gamma_t(\mathbf{x}_0\mid \mathbf{x}_t)}
    \left[
        v_t(\mathbf{x}_t\mid \mathbf{x}_0)
        \right],
    \label{eq:app-vfm-posterior-average}
\end{equation}
where $\Gamma_t(\mathbf{x}_0\mid\mathbf{x}_t)$ is the posterior over source endpoints induced by the interpolant.
Thus learning the marginal vector field can be viewed as approximating this endpoint posterior.

Variational flow matching introduces a recognition model
$q_{t,\phi}(\mathbf{x}_0\mid\mathbf{x}_t)$ and minimizes
\begin{equation}
    \mathcal{L}_{\mathrm{VFM}}(\phi)
    =
    \mathbb{E}_{t,\mathbf{x}_t}
    \left[
        \mathrm{KL}\!\left(
        \Gamma_t(\cdot\mid\mathbf{x}_t)
        \,\|\,
        q_{t,\phi}(\cdot\mid\mathbf{x}_t)
        \right)
        \right].
    \label{eq:app-vfm-kl}
\end{equation}
Equivalently, up to a constant independent of $\phi$,
\begin{equation}
    \mathcal{L}_{\mathrm{VFM}}(\phi)
    \equiv
    -
    \mathbb{E}_{t,\mathbf{x}_t,\mathbf{x}_0}
    \left[
        \log q_{t,\phi}(\mathbf{x}_0\mid\mathbf{x}_t)
        \right],
    \qquad
    \mathbf{x}_0\sim \Gamma_t(\cdot\mid\mathbf{x}_t).
    \label{eq:app-vfm-nll}
\end{equation}
This is the variational inference view: flow matching is recast as posterior approximation over the endpoint variable that generated the intermediate state.

Now assume a fixed-covariance Gaussian recognition family,
\[
    q_{t,\phi}(\mathbf{x}_0\mid\mathbf{x}_t)
    =
    \mathcal{N}\bigl(
    \mu_\phi(\mathbf{x}_t,t),
    \sigma_{\mathbf{x}_0}^2 I
    \bigr).
\]
Then Eq.~\eqref{eq:app-vfm-nll} is equivalent, up to constants and a positive scale, to regression of the posterior mean:
\begin{equation}
    \mathcal{L}_{\mathrm{VFM}}(\phi)
    \equiv
    \mathbb{E}
    \left[
        \left\|
        \mu_\phi(\mathbf{x}_t,t)-\mathbf{x}_0
        \right\|^2
        \right].
    \label{eq:app-vfm-mean-regression}
\end{equation}
Since Eq.~\eqref{eq:app-vfm-cond-velocity} is affine in $\mathbf{x}_0$, posterior-mean regression can be written as a time-weighted velocity regression with the induced vector field
\begin{equation}
    v_{\phi,t}(\mathbf{x}_t)
    =
    \frac{\partial_t f(t)}{1-f(t)}
    \bigl(
    \mu_\phi(\mathbf{x}_t,t)-\mathbf{x}_t
    \bigr),
    \label{eq:app-vfm-induced-velocity}
\end{equation}
Indeed,
\begin{equation}
    \mathcal{L}_{\mathrm{VFM}}(\phi)
    \equiv
    \mathbb{E}_{t,\mathbf{x}_0,\mathbf{x}_1}
    \left[
        \left(
        \frac{1-f(t)}{\partial_t f(t)}
        \right)^2
        \left\|
        v_{\phi,t}(\mathbf{x}_t)
        -
        v_t(\mathbf{x}_t\mid\mathbf{x}_0)
        \right\|^2
        \right].
    \label{eq:app-vfm-velocity-regression}
\end{equation}
Thus, Gaussian VFM and standard flow matching use the same conditional velocity target, differing only by a time-dependent weighting of the regression loss.
Under the linear schedule $f(t)=1-t$, this recovers the simulation-free flow-matching objective used by SCFM.
The distinction is that SCFM applies this posterior matching to the structured source endpoint $\mathbf{x}_0=(\mathbf{z},\varepsilon)$ rather than to an unstructured noise variable.

\section{SCFM: Extended Derivations and Implementation Details}
\label{app:scfm}

This appendix complements Section~\ref{sec:method-scfm} with extended derivations and implementation details. Specifically, it gives the endpoint derivation, the endpoint consistency argument, the intermediate-time specialization to VFM, and the practical SCFM objective.

\paragraph{Shared posterior-matching view.}
We extend the posterior-matching construction (see Appendix~\ref{app:vfm-view}), such that the source endpoint becomes structured as $\mathbf{x}_0=(\mathbf{z},\varepsilon)$. The remainder of this appendix focuses on the two SCFM-specific regimes: the endpoint derivation at $t=1$ and the intermediate-time specialization for $t<1$.

\subsection{Endpoint regime: posterior identity and endpoint objective}
\label{app:scfm-endpoint-objective}

At $t=1$, the interpolant endpoint is $\mathbf{x}_t=\mathbf{x}_1$. Under the encoder-induced coupling in Eq.~\eqref{eq:structured_coupling}, the source posterior becomes
\[
    \Gamma_1(\mathbf{x}_0\mid\mathbf{x}_1)
    =
    q_\phi(\mathbf{z}\mid\mathbf{x}_1)\,p(\varepsilon),
    \qquad
    \mathbf{x}_0=(\mathbf{z},\varepsilon).
\]
Using the SCFM decoder factorization
\[
    p_{\theta,\psi}(\mathbf{x}_1,\mathbf{x}_0)
    =
    p_\theta(\mathbf{x}_1\mid\mathbf{z})\,p_\psi(\mathbf{z})\,p(\varepsilon),
\]
the shared posterior mean splits as
\[
    \mu_\phi(\mathbf{x}_1,1)
    =
    \bigl(
    \mu_\phi^z(\mathbf{x}_1),
    \mu_\phi^\varepsilon(\mathbf{x}_1)
    \bigr),
\]
with latent posterior
\[
    q_\phi(\mathbf{z}\mid\mathbf{x}_1)
    =
    \mathcal{N}\bigl(
    \mathbf{z};\,\mu_\phi^z(\mathbf{x}_1),
    \operatorname{diag}(\sigma_\phi^2(\mathbf{x}_1))
    \bigr).
\]
The endpoint mean is recovered from the shared velocity parameterization through
\begin{equation}
    \mu_\phi(\mathbf{x}_1,1)
    =
    \mathbf{x}_1
    +
    \frac{1}{\partial_t f(1)}\,v_{\phi,1}(\mathbf{x}_1).
    \label{eq:mu-from-velocity}
\end{equation}

For compactness, write
\[
    q_x(\mathbf{z}) := q_\phi(\mathbf{z}\mid\mathbf{x}_1),
    \qquad
    q_x^\varepsilon(\varepsilon)
    :=
    \mathcal{N}\bigl(
    \varepsilon;\,
    \mu_\phi^\varepsilon(\mathbf{x}_1),
    I_{d_\varepsilon}
    \bigr).
\]
Because the decoder depends only on $\mathbf{z}$, \(p_\theta(\mathbf{x}_1\mid\mathbf{x}_0)=p_\theta(\mathbf{x}_1\mid\mathbf{z})\). The endpoint KL then decomposes as
\begin{align}
    \mathrm{KL}\big(
    q_x\,q_x^\varepsilon
    \,\|\,
    p_\psi(\mathbf{z})\,p(\varepsilon)
    \big)
     & =
    \mathrm{KL}\big(
    q_x
    \,\|\,
    p_\psi(\mathbf{z})
    \big)
    \notag     \\
     & \quad +
    \mathrm{KL}\big(
    q_x^\varepsilon
    \,\|\,
    p(\varepsilon)
    \big)
    \notag     \\
     & =
    \mathrm{KL}\big(
    q_x
    \,\|\,
    p_\psi(\mathbf{z})
    \big)
    +
    \frac{1}{2}\|\mu_\phi^\varepsilon(\mathbf{x}_1)\|^2.
\end{align}
Likewise,
\[
    \mathbb{E}_{q_x q_x^\varepsilon}
    \big[
        \log p_\theta(\mathbf{x}_1\mid\mathbf{x}_0)
        \big]
    =
    \mathbb{E}_{q_x}
    \big[
        \log p_\theta(\mathbf{x}_1\mid\mathbf{z})
        \big].
\]
Therefore the endpoint contribution is
\begin{equation}
    \begin{aligned}
        \mathcal{L}_{\mathrm{end}}
         & =
        \mathbb{E}_{p_{\text{data}}(\mathbf{x}_1)}
        \Big[
            -\,\mathbb{E}_{q_x}
            \log p_\theta(\mathbf{x}_1\mid\mathbf{z})
            +
            \mathrm{KL}\bigl(
            q_x
            \,\|\,
            p_\psi(\mathbf{z})
            \bigr)
            +
            \frac{1}{2}\|\mu_\phi^\varepsilon(\mathbf{x}_1)\|^2
            \Big]
        \\
         & =
        \mathcal{L}_{\mathrm{VAE}}
        +
        \mathcal{R}_{\varepsilon}.
    \end{aligned}
    \label{eq:app-endpoint-objective}
\end{equation}
This is exactly the endpoint decomposition stated in Section~\ref{sec:method-scfm}, with $\mathcal{R}_{\varepsilon}$ matching the quadratic penalty in Eq.~\eqref{eq:exogenous_regularizer}.

\subsection{Coupling consistency and prior matching}
\label{app:scfm-endpoint-consistency}

For compactness, define the encoder-induced and decoder-induced latent joints
\[
    \Gamma_\phi^{\mathrm{enc}}(\mathbf{x}_1,\mathbf{z})
    =
    p_{\text{data}}(\mathbf{x}_1)\,q_\phi(\mathbf{z}\mid\mathbf{x}_1)
\]
\[
    \Gamma_{\theta,\psi}^{\mathrm{dec}}(\mathbf{x}_1,\mathbf{z})
    =
    p_\psi(\mathbf{z})\,p_\theta(\mathbf{x}_1\mid\mathbf{z}).
\]
Then
\[
    \mathrm{KL}\bigl(
    \Gamma_\phi^{\mathrm{enc}}
    \,\|\,
    \Gamma_{\theta,\psi}^{\mathrm{dec}}
    \bigr)
    =
    \mathcal{L}_{\mathrm{VAE}}
    +
    \mathrm{const},
\]
where the constant depends only on \(p_{\text{data}}\).
As in Section~\ref{sec:method-augmented-source}, the exact identity
\[
    q_\phi^{\mathrm{agg}}(\mathbf{z})=p_\psi(\mathbf{z})
\]
is a global-optimum statement that relies on sufficiently expressive encoder, decoder, and prior families \citep{hoffman2016elbo,pmlr-v80-alemi18a}. In practice, the more relevant statement is that the endpoint term controls the prior--aggregated-posterior mismatch through an upper bound. Let \(T(\mathbf{x}_1,\mathbf{z})=\mathbf{z}\) be the projection onto the latent coordinate. Then
\[
    T_\#\Gamma_\phi^{\mathrm{enc}}
    =
    q_\phi^{\mathrm{agg}}(\mathbf{z}),
    \qquad
    T_\#\Gamma_{\theta,\psi}^{\mathrm{dec}}
    =
    p_\psi(\mathbf{z}),
\]
so the KL data-processing inequality under marginalization gives
\begin{equation}
    \mathrm{KL}\bigl(
    q_\phi^{\mathrm{agg}}(\mathbf{z})
    \,\|\,
    p_\psi(\mathbf{z})
    \bigr)
    \le
    \mathrm{KL}\bigl(
    \Gamma_\phi^{\mathrm{enc}}
    \,\|\,
    \Gamma_{\theta,\psi}^{\mathrm{dec}}
    \bigr)
    =
    \mathcal{L}_{\mathrm{VAE}}
    +
    \mathrm{const}.
    \label{eq:app-agg-bound}
\end{equation}
Consequently, reducing the endpoint VAE objective also reduces an upper bound on the mismatch between the encoder-induced training marginal and the sampling prior. If this joint KL is zero, then
\[
    \Gamma_\phi^{\mathrm{enc}}(\mathbf{x}_1,\mathbf{z})
    =
    \Gamma_{\theta,\psi}^{\mathrm{dec}}(\mathbf{x}_1,\mathbf{z})
    \qquad
    \text{a.e.}
\]
and marginalizing over $\mathbf{x}_1$ yields
\[
    q_\phi^{\mathrm{agg}}(\mathbf{z})
    :=
    \int p_{\text{data}}(\mathbf{x}_1)\,
    q_\phi(\mathbf{z}\mid\mathbf{x}_1)\,d\mathbf{x}_1
    =
    p_\psi(\mathbf{z}).
\]
This is the endpoint consistency statement used in the main text: exact endpoint alignment recovers the same source marginal that is used at sampling time.

\subsection{Intermediate-time regime: specialization to VFM}
\label{app:scfm-intermediate-vfm}

For $t\in[0,1)$, SCFM reduces to the same posterior-matching problem analyzed in Appendix~\ref{app:vfm-view}, now with structured endpoint variable $\mathbf{x}_0=(\mathbf{z},\varepsilon)$. We use the shared Gaussian family
\begin{equation}
    q_{t,\phi}(\mathbf{x}_0\mid\mathbf{x}_t)
    =
    \mathcal{N}\bigl(
    \mu_\phi(\mathbf{x}_t,t),
    \sigma_{\mathbf{x}_0}^2 I_D
    \bigr),
    \label{eq:qt-gaussian}
\end{equation}
whose mean is shared with the endpoint encoder, while the endpoint covariance of $q_\phi(\mathbf{z}\mid\mathbf{x}_1)$ is predicted by a separate variance head. Under the linear interpolant,
\[
    \mathbf{x}_t
    =
    f(t)\mathbf{x}_0+(1-f(t))\mathbf{x}_1,
\]
the conditional velocity is
\begin{equation}
    v_t(\mathbf{x}_t\mid\mathbf{x}_0)
    =
    \frac{\partial_t f(t)}{1-f(t)}
    \bigl(
    \mathbf{x}_0-\mathbf{x}_t
    \bigr).
    \label{eq:app-scfm-cond-velocity}
\end{equation}
The corresponding variational objective is
\begin{equation}
    \mathcal{L}_{\mathrm{VFM}}(\phi)
    =
    \mathbb{E}_{t\sim\mathcal{U}[0,1),\,\mathbf{x}_t\sim p_t}
    \Big[
        \mathrm{KL}\big(
        \Gamma_t(\cdot\mid\mathbf{x}_t)
        \,\|\,
        q_{t,\phi}(\cdot\mid\mathbf{x}_t)
        \big)
        \Big].
    \label{eq:app-scfm-vfm-kl}
\end{equation}
When training $q_{t,\phi}$, we use the equivalent stop-gradient form
\begin{equation}
    \mathcal{L}_{\mathrm{VFM}}(\phi)
    \equiv
    -\,
    \mathbb{E}_{t,(\mathbf{x}_t,\mathbf{x}_0)\sim\Gamma_t}
    \Big[
        \log q_{t,\phi}\bigl(
        \operatorname{sg}(\mathbf{x}_0)
        \mid
        \operatorname{sg}(\mathbf{x}_t)
        \bigr)
        \Big]
    + \mathrm{const.},
    \label{eq:app-scfm-vfm-stopgrad}
\end{equation}
where, in practice, \(\mathbf{x}_1\sim p_{\mathrm{data}}\), \(\mathbf{z}\sim q_\phi(\mathbf{z}\mid\mathbf{x}_1)\), \(\varepsilon\sim p(\varepsilon)\), \(\mathbf{x}_0=(\operatorname{sg}(\mathbf{z}),\varepsilon)\), and \(\mathbf{x}_t=f(t)\mathbf{x}_0+(1-f(t))\mathbf{x}_1\).
Under the Gaussian family in Eq.~\eqref{eq:qt-gaussian}, Appendix~\ref{app:vfm-view} implies that minimizing Eq.~\eqref{eq:app-scfm-vfm-kl} is equivalent to posterior-mean regression and therefore to the time-weighted velocity regression
\begin{align}
    \mathcal{L}_{\mathrm{VFM}}(\phi)
     & \equiv
    \mathbb{E}_{t,\mathbf{x}_0,\mathbf{x}_1}
    \left[
        \left(
        \frac{1-f(t)}{\partial_t f(t)}
        \right)^2
        \left\|
        v_{\phi,t}(\mathbf{x}_t)
        -
        v_t(\mathbf{x}_t\mid\mathbf{x}_0)
        \right\|^2
        \right],
    \notag          \\
     & \hspace{4em}
    \mathbf{x}_t=f(t)\mathbf{x}_0+(1-f(t))\mathbf{x}_1.
    \label{eq:app-scfm-vfm-regression}
\end{align}

\subsection{Practical SCFM Objective and Prior Parameterization}
\label{app:scfm-practical-training}

The practical SCFM objective combines the intermediate-time VFM loss with an endpoint latent-variable objective:
\[
    \mathcal{L}_{\mathrm{SCFM}}
    =
    \mathcal{L}_{\mathrm{VFM}}
    +
    \mathcal{L}_{\mathrm{end}}^{\mathrm{train}}.
\]
The endpoint term is implemented as
\[
    \mathcal{L}_{\mathrm{end}}^{\mathrm{train}}
    =
    \mathcal{L}_{\mathrm{rec}}
    +
    \mathcal{R}_{z}
    +
    \mathcal{R}_{\varepsilon},
    \qquad
    \mathcal{L}_{\mathrm{rec}}
    =
    \mathcal{L}_{\mathrm{pix}}
    +
    \alpha_{\mathrm{perc}}\mathcal{L}_{\mathrm{perc}}.
\]
The perceptual term can be instantiated with LPIPS~\citep{zhang2018unreasonable}, following the use of perceptual reconstruction losses in image generation~\citep{johnson2016perceptual}. The latent regularizer $\mathcal{R}_z$ is instantiated as either a $\beta$-VAE penalty~\citep{higgins2017betavae},
\[
    \mathcal{R}_{z}^{\beta\text{-VAE}}
    =
    \beta\,\mathrm{KL}
    \bigl(
    q_\phi(\mathbf{z}\mid\mathbf{x}_1)
    \,\|\,
    p_\psi(\mathbf{z})
    \bigr),
\]
or a $\beta$-TCVAE-style regularizer~\citep{chen2018isolating},
\[
    \mathcal{R}_{z}^{\beta\text{-TCVAE}}
    =
    \mathrm{KL}
    \bigl(
    q_\phi(\mathbf{z}\mid\mathbf{x}_1)
    \,\|\,
    p_\psi(\mathbf{z})
    \bigr)
    +
    \beta\,\mathrm{TC}
    \bigl(q_\phi^{\mathrm{agg}}(\mathbf{z})\bigr),
\]
where \(q_\phi^{\mathrm{agg}}(\mathbf{z})=\int p_{\mathrm{data}}(\mathbf{x}_1)q_\phi(\mathbf{z}\mid\mathbf{x})\,d\mathbf{x}\) and \(\mathrm{TC}(q)=\mathrm{KL}\bigl(q(\mathbf{z})\,\|\,\prod_j q(z_j)\bigr)\).
Thus, $\beta$-VAE and $\beta$-TCVAE provide two endpoint regularizers for structured latent learning under the same prior. These choices affect only the endpoint term and leave $\mathcal{L}_{\mathrm{VFM}}$ unchanged.

\paragraph{GMM prior.}
We parameterize the structured prior as a learnable Gaussian mixture,
\[
    p_\psi(\mathbf{z})
    =
    \sum_{k=1}^{K}
    \pi_k\,
    \mathcal{N}
    \bigl(
    \mathbf{z};
    \mu_{\psi,k},
    \operatorname{diag}(\sigma^2_{\psi,k})
    \bigr),
    \qquad
    \sum_{k=1}^{K}\pi_k=1.
\]
The KL term is estimated by sampling
\(
\mathbf{z}
=
\mu_\phi^z(\mathbf{x})
+
\sigma_\phi(\mathbf{x})\odot\boldsymbol{\xi}
\),
with
\(
\boldsymbol{\xi}\sim\mathcal{N}(0,I)
\),
and evaluating
\(
\mathrm{KL}(q_\phi(\mathbf{z}\mid\mathbf{x}_1)\|p_\psi(\mathbf{z}))
=
\mathbb{E}_{q_\phi}
[
\log q_\phi(\mathbf{z}\mid\mathbf{x}_1)
-
\log p_\psi(\mathbf{z})
]
\).
At sampling time, we draw
\(
\mathbf{z}\sim p_\psi(\mathbf{z})
\)
and
\(
\varepsilon\sim\mathcal{N}(0,I)
\),
forming the structured source
\(
\mathbf{x}_0=(\mathbf{z},\varepsilon)
\).

\section{SCFM Algorithms}
\label{app:algorithms}
This section collects compact pseudocode summaries of the practical SCFM procedures described in Sections~\ref{sec:method-objective} and~\ref{sec:method-sampling}.

\subsection{Training}
Algorithm~\ref{alg:scfm_training} summarizes one SCFM training step, including the endpoint latent-variable terms and the intermediate-time flow-matching update.
\begin{algorithm}[!htbp]
    \small
    \caption{One SCFM training step}
    \label{alg:scfm_training}
    \begin{algorithmic}[1]
        \Require minibatch $\mathbf{x}_1 \sim p_{\mathrm{data}}$, velocity model $v_{\phi,t}$, endpoint variance head $\sigma_\phi^2(\mathbf{x}_1)$, decoder $p_\theta(\mathbf{x}_1\mid\mathbf{z})$, latent prior $p_\psi(\mathbf{z})$
        \State \textit{Endpoint VAE regime:}
        \State $\quad$ Compute $\mu_\phi(\mathbf{x}_1,1)=\bigl(\mu_\phi^z(\mathbf{x}_1),\mu_\phi^\varepsilon(\mathbf{x}_1)\bigr)$
        \State $\quad$ Form $q_\phi(\mathbf{z}\mid\mathbf{x}_1)=\mathcal{N}\bigl(\mu_\phi^z(\mathbf{x}_1),\operatorname{diag}(\sigma_\phi^2(\mathbf{x}_1))\bigr)$
        \State $\quad$ Sample $\mathbf{z}\sim q_\phi(\mathbf{z}\mid\mathbf{x}_1)$ and compute $\mathcal{L}_{\mathrm{rec}}$ from $p_\theta(\mathbf{x}_1\mid\mathbf{z})$
        \State $\quad$ Compute $\mathcal{L}_{\mathrm{KL}}=\mathrm{KL}\bigl(q_\phi(\mathbf{z}\mid\mathbf{x}_1)\,\|\,p_\psi(\mathbf{z})\bigr)$
        \State $\quad$ Compute $\mathcal{R}_{\varepsilon}=\frac{1}{2}\|\mu_\phi^\varepsilon(\mathbf{x}_1)\|^2$
        \State \textit{Flow-matching regime:}
        \State $\quad$ Sample $\varepsilon\sim p(\varepsilon)=\mathcal{N}(0,I_{d_\varepsilon})$ and set $\mathbf{x}_0=\bigl(\operatorname{sg}(\mathbf{z}),\varepsilon\bigr)$
        \State $\quad$ Sample $t\sim \mathcal{U}[0,1)$ and form $\mathbf{x}_t=f(t)\mathbf{x}_0+(1-f(t))\mathbf{x}_1$ with $f(t)=1-t$
        \State $\quad$ Compute
        $\mathcal{L}_{\mathrm{VFM}}
            =
            \left(\frac{1-f(t)}{\partial_t f(t)}\right)^2
            \left\|
            v_{\phi,t}(\mathbf{x}_t)
            -
            \frac{\partial_t f(t)}{1-f(t)}
            \bigl(\mathbf{x}_0-\mathbf{x}_t\bigr)
            \right\|^2$
        \State Form $\mathcal{L}_{\mathrm{SCFM}}=\mathcal{L}_{\mathrm{VFM}}+\mathcal{L}_{\mathrm{rec}}+\mathcal{L}_{\mathrm{KL}}+\mathcal{R}_{\varepsilon}$
        \State Update $(\theta,\phi,\psi)$ using $\mathcal{L}_{\mathrm{SCFM}}$
    \end{algorithmic}
\end{algorithm}

\subsection{Sampling}
Algorithm~\ref{alg:scfm_sampling} gives the full ODE sampler, while Algorithm~\ref{alg:scfm_fast_sampling} summarizes the decoder-initialized refinement sampler that initializes near the data manifold using the decoder.
\begin{algorithm}[!htbp]
    \small
    \caption{Full ODE SCFM sampling / reconstruction}
    \label{alg:scfm_sampling}
    \begin{algorithmic}[1]
        \Require velocity model $v_{\phi,t}$, endpoint variance head $\sigma_\phi^2(\mathbf{x}_1)$, latent prior $p_\psi(\mathbf{z})$, optional observation $\mathbf{x}_1$
        \If{an observation $\mathbf{x}_1$ is available}
        \State Compute $\mu_\phi(\mathbf{x}_1,1)=\bigl(\mu_\phi^z(\mathbf{x}_1),\mu_\phi^\varepsilon(\mathbf{x}_1)\bigr)$
        \State Form $q_\phi(\mathbf{z}\mid\mathbf{x}_1)=\mathcal{N}\bigl(\mu_\phi^z(\mathbf{x}_1),\operatorname{diag}(\sigma_\phi^2(\mathbf{x}_1))\bigr)$
        \State Sample $\mathbf{z}\sim q_\phi(\mathbf{z}\mid\mathbf{x}_1)$
        \Else
        \State Sample $\mathbf{z}\sim p_\psi(\mathbf{z})$
        \EndIf
        \State Sample $\varepsilon\sim p(\varepsilon)=\mathcal{N}(0,I_{d_\varepsilon})$ and set $\mathbf{x}_0=(\mathbf{z},\varepsilon)$
        \State Integrate the ODE with velocity field $v_{\phi, t}(\mathbf{x}_t)$ on $t\in[0,1]$ with initial condition $\mathbf{x}_{t=0}=\mathbf{x}_0$
        \State \Return $\mathbf{x}_{t=1}$
    \end{algorithmic}
\end{algorithm}

\begin{algorithm}[!htbp]
    \small
    \caption{Decoder-initialized refinement SCFM sampling / reconstruction}
    \label{alg:scfm_fast_sampling}
    \begin{algorithmic}[1]
        \Require start time $t_0\in(0,1)$, velocity model $v_{\phi,t}$, endpoint variance head $\sigma_\phi^2(\mathbf{x}_1)$, decoder $p_\theta(\mathbf{x}_1\mid\mathbf{z})$, latent prior $p_\psi(\mathbf{z})$, optional observation $\mathbf{x}_1$
        \If{an observation $\mathbf{x}_1$ is available}
        \State Compute $\mu_\phi(\mathbf{x}_1,1)=\bigl(\mu_\phi^z(\mathbf{x}_1),\mu_\phi^\varepsilon(\mathbf{x}_1)\bigr)$
        \State Form $q_\phi(\mathbf{z}\mid\mathbf{x}_1)=\mathcal{N}\bigl(\mu_\phi^z(\mathbf{x}_1),\operatorname{diag}(\sigma_\phi^2(\mathbf{x}_1))\bigr)$
        \State Sample $\mathbf{z}\sim q_\phi(\mathbf{z}\mid\mathbf{x}_1)$
        \Else
        \State Sample $\mathbf{z}\sim p_\psi(\mathbf{z})$
        \EndIf
        \State Sample $\varepsilon\sim p(\varepsilon)=\mathcal{N}(0,I_{d_\varepsilon})$ and set $\mathbf{x}_0=(\mathbf{z},\varepsilon)$
        \State Sample $\widehat{\mathbf{x}}_1\sim p_\theta(\cdot\mid\mathbf{z})$
        \State Form $\mathbf{x}_{t_0}=f(t_0)\mathbf{x}_0+(1-f(t_0))\widehat{\mathbf{x}}_1$ with $f(t)=1-t$
        \State Integrate the ODE with velocity field $v_{\phi, t}(\mathbf{x}_t)$ on $t\in[t_0,1]$ with initial condition $\mathbf{x}_{t=t_0}=\mathbf{x}_{t_0}$
        \State \Return $\mathbf{x}_{t=1}$
    \end{algorithmic}
\end{algorithm}

\section{Extended Related Work}
\label{app:related-work}

\paragraph{Structured priors and couplings.}
Several lines of work mitigate limitations of fixed, unstructured priors in diffusion- and transport-based models.
PriorGrad \citep{lee2022priorgrad} learns a data-dependent Gaussian prior for conditional diffusion, while DecompDiff \citep{10.5555/3618408.3618882} uses decomposed priors for structured molecule generation.
Other approaches leave the prior fixed but change the pairing between source and data samples, for example through minibatch optimal transport \citep{tong2024improving}, data-dependent couplings in stochastic interpolants \citep{pmlr-v235-albergo24a}, or post-hoc correction of prior mismatch with optimal transport maps \citep{2024arXiv241013431W}.
These methods improve the transport problem or sampling geometry, but generally do not make the source itself a learned representation space.

\paragraph{Variational coupling.}
Variational Consistency Training \citep{silvestri2025vct} is closest to our use of an encoder for coupling: it learns a data-noise coupling with a VAE-style encoder and a KL penalty that keeps emitted noise close to the sampling prior, reducing variance in consistency training.
SCFM uses a related variational mechanism for a different purpose.
Rather than learning a coupling only to stabilize a consistency objective, we use the encoder to define a structured source posterior whose latent component is trained as a representation and whose marginal prior is learned jointly with the flow.

\paragraph{Flow matching for representations.}
Latent-variable formulations of flow and rectified-flow matching model transport ambiguity with hidden variables \citep{guo2025variational,zhang2025towards}.
Structured Flow Autoencoders \citep{xu2026structured} are especially close in motivation: they use posterior-averaged conditional vector fields to combine flow matching with structured probabilistic representations.
Recent concurrent work, FlowFM \citep{ukita2026highperformance}, jointly trains a representation encoder and a conditional flow-matching generator for self-supervised learning, emphasizing downstream recognition performance and computational efficiency on wearable-sensor data.
Recent work on disentangled representation learning via flow matching \citep{chi2026disentangled} learns factor-conditioned latent-space flows and regularizes factor-specific velocity components to reduce cross-factor leakage.
SCFM differs from these lines by learning a structured \emph{source distribution} for a stochastic interpolant: the latent variable is part of the source endpoint being transported, not only a conditioning variable for a generator or a factorization of the velocity field.

\paragraph{Latent-variable models.}
Variational autoencoders and their extensions learn latent representations through amortized inference and variational training \citep{Kingma2014,burda2015importance}.
This family includes structured-prior models for clustering, such as Gaussian-mixture and multi-facet variants \citep{dilokthanakul2016deep,jiang2017variational,falck2021multi}, as well as regularized objectives that encourage disentangled latent factors \citep{higgins2017betavae,burgess2018understanding,chen2018isolating,pmlrv97locatello19a}.
These methods show that explicit latent variables are effective for representation learning, but their sample quality is often limited by the decoder family and by the fact that generation is tied directly to the latent-variable model.
SCFM inherits the representation-learning advantages of this line of work, but differs in how generation is performed: the latent model is used to learn a structured source and endpoint supervision, while the main sampler remains a flow that transports samples from the learned source prior to the data distribution.

\section{Experimental Details}
\label{app:experimental_details}
This appendix provides the implementation details needed to reproduce the experiments in Section~\ref{sec:experiments}. We describe the dataset-specific architectures, training schedules, latent dimensionalities, endpoint objectives, priors, samplers, and probing protocols used for MNIST, CIFAR-10, Cars3D, Shapes3D, and ImageNet-128. Unless otherwise stated, all models are trained without using class or factor labels; labels and ground-truth factors are used only for downstream probing or disentanglement evaluation.

Experiments were run on a heterogeneous pool of GPUs: 2 NVIDIA RTX PRO 6000 Blackwell Server Edition GPUs, 2 NVIDIA A100 80GB PCIe GPUs, 2 NVIDIA RTX A6000 GPUs, and 3 Quadro RTX 6000 GPUs.

For the Flow Matching baselines on CIFAR-10 and ImageNet-128, we use the same encoder backbone and training configuration as SCFM, but remove the structured endpoint branch. In particular, the baseline is trained only with the standard flow-matching regression objective and does not include the additional latent head, GMM prior, or VAE-style endpoint loss. This ensures that differences between FM and SCFM are attributable to the structured coupling and endpoint objective rather than changes in architecture or optimization.

\subsection{MNIST Clustering Setup}
\label{app:mnist_setup}

Across all methods, we set the number of mixture components to $K=10$ to match the ten MNIST classes and to keep the comparison aligned across models. Reported metrics are computed over five independent runs with different random seeds, and we report the mean and standard deviation across runs.

\begin{table}[!htbp]
    \caption{\textbf{MNIST clustering settings.}}
    \label{tab:mnist_settings}
    \centering
    \footnotesize
    \setlength{\tabcolsep}{4pt}
    \renewcommand{\arraystretch}{1.05}
    \begin{tabular}{lcccc}
        \toprule
         & \textbf{VaDE}               & \textbf{MFCVAE} & \textbf{SCFM ($\beta$-VAE)} & \textbf{SCFM ($\beta$-TCVAE)} \\
        \midrule
        Latent
         & $z\in\mathbb{R}^{10}$
         & $J{=}2,\ z_j{=}5$
         & $z\in\mathbb{R}^{10}$
         & $z\in\mathbb{R}^{10}$                                                                                       \\
        Components
         & $K{=}10$
         & $K_j{=}10$
         & $K{=}10$
         & $K{=}10$                                                                                                    \\
        Objective
         & $\beta{=}1$ + GMM
         & \makecell[c]{MFCVAE + diag.                                                                                 \\$p(z_j\mid c_j)$}
         & GMM + $\beta{=}4$
         & GMM + $\beta{=}3$ (TC)                                                                                      \\
        Batch size
         & $256$                       & $256$           & $256$                       & $256$                         \\
        Optimizer
         & AdamW ($10^{-3}$)
         & Adam ($5{\times}10^{-4}$)
         & AdamW ($10^{-3}$)
         & AdamW ($10^{-3}$)                                                                                           \\
        \bottomrule
    \end{tabular}
\end{table}

\subsection{CIFAR-10 Setup}
\label{app:cifar10_setup}

Table~\ref{tab:cifar10_scfm_arch_training} summarizes the encoder, decoder, training, and sampling settings used for the CIFAR-10 SCFM experiments.

\begin{table}[!htbp]
    \centering
    \caption{\textbf{SCFM CIFAR-10 network and training setup.}
        The encoder parameterizes the shared posterior/recognition network, while the decoder is used for the endpoint reconstruction objective and decoder-initialized refinement.}
    \label{tab:cifar10_scfm_arch_training}
    \small
    \setlength{\tabcolsep}{4pt}
    \renewcommand{\arraystretch}{1.05}
    \begin{tabular}{lcc}
        \toprule
        \textbf{Hyperparameter} & \textbf{Encoder / recognition net}                                              & \textbf{Decoder} \\
        \midrule
        Role
                                & \makecell[c]{Shared $q_{t,\phi}(x_0\mid x_t)$                                                      \\and endpoint encoder $q_\phi(z\mid x)$}
                                & \makecell[c]{Endpoint decoder                                                                      \\$p_\theta(x\mid z)$} \\
        Backbone
                                & U-Net
                                & U-Net                                                                                              \\
        Channels
                                & $128$
                                & $128$                                                                                              \\
        Depth
                                & $4$
                                & $4$                                                                                                \\
        Channel multipliers
                                & $(2,2,2)$
                                & $(2,2,2)$                                                                                          \\
        Heads
                                & $1$
                                & --                                                                                                 \\
        Head channels
                                & $-1$
                                & --                                                                                                 \\
        Attention resolution
                                & $[2]$
                                & --                                                                                                 \\
        Dropout
                                & $0.3$
                                & --                                                                                                 \\
        Latent dimension
                                & \multicolumn{2}{c}{$64$}                                                                           \\
        Training length
                                & \multicolumn{2}{c}{$1300$ epochs, $\approx 508$K iterations}                                       \\
        Optimizer
                                & \multicolumn{2}{c}{Adam, $\beta_1=0.9$, $\beta_2=0.999$, $\varepsilon=10^{-8}$}                    \\
        Learning rate
                                & \multicolumn{2}{c}{$5{\times}10^{-4}$ with polynomial decay}                                       \\
        Warmup
                                & \multicolumn{2}{c}{$45$K steps, linear warmup from $10^{-8}$}                                      \\
        Structured prior
                                & \multicolumn{2}{c}{Learnable GMM prior $p_\psi(z)$ with $K=10$ components}                         \\
        Reconstruction loss
                                & \multicolumn{2}{c}{Pixel reconstruction with LPIPS augmentation}                                   \\
        Endpoint objective
                                & \multicolumn{2}{c}{$\beta$-VAE with $\beta=2.8$}                                                   \\
        EMA
                                & \multicolumn{2}{c}{Enabled, decay $0.9999$}                                                        \\
        Sampler
                                & \multicolumn{2}{c}{Heun ODE sampler}                                                               \\
        \bottomrule
    \end{tabular}
\end{table}

\subsection{Cars3D and Shapes3D Disentanglement Setup}
\label{app:cars3d_shapes3d_setup}
Table~\ref{tab:cars3d_shapes3d_setup} summarizes the dataset, training. We follow the standard disentanglement benchmark protocol of \citet{pmlrv97locatello19a}: models are trained without access to ground-truth factor labels, while factor annotations are used only for evaluation. For each dataset and endpoint regularizer, we train models across $10$ random seeds and sweep over the corresponding $\beta$ values for the VAE-style objective. All models use convolutional encoder--decoder networks and a $10$-dimensional latent representation.
\begin{table}[!htbp]
    \centering
    \caption{\textbf{Cars3D and Shapes3D disentanglement setup.}
        Models are trained without factor labels; ground-truth factors are used only for disentanglement evaluation.}
    \label{tab:cars3d_shapes3d_setup}
    \small
    \setlength{\tabcolsep}{4pt}
    \renewcommand{\arraystretch}{1.05}
    \begin{tabular}{lcc}
        \toprule
         & \textbf{Cars3D}                                                                  & \textbf{Shapes3D} \\
        \midrule
        Resolution
         & $64{\times}64{\times}3$
         & $64{\times}64{\times}3$                                                                              \\
        Images
         & $17{,}568$
         & $480{,}000$                                                                                          \\
        Factors
         & $3$
         & $6$                                                                                                  \\
        Cardinalities
         & $183{\times}4{\times}24$
         & $10{\times}10{\times}10{\times}8{\times}4{\times}15$                                                 \\
        Factor names
         & \makecell[c]{car ID, elevation,                                                                      \\azimuth}
         & \makecell[c]{floor hue, wall hue, object hue,                                                        \\scale, shape, orientation} \\
        Train labels
         & None
         & None                                                                                                 \\
        Eval factors
         & Used for metrics only
         & Used for metrics only                                                                                \\
        Latent dimension
         & $10$
         & $10$                                                                                                 \\
        Representation
         & $\mu_\phi(x)$
         & $\mu_\phi(x)$                                                                                        \\
        $\beta$-VAE sweep
         & $\{1,2,4,6,8,16\}$
         & $\{1,2,4,6,8,16\}$                                                                                   \\
        $\beta$-TCVAE sweep
         & $\{1,2,4,6,8,10\}$
         & $\{1,2,4,6,8,10\}$                                                                                   \\
        Training
         & \multicolumn{2}{c}{Adam, learning rate $10^{-4}$, batch size $64$, $300$K steps}                     \\
        \bottomrule
    \end{tabular}
\end{table}

\subsection{ImageNet-128 Setup}
\label{app:imagenet128_setup}

\begin{figure}[!htbp]
    \centering
    \makebox[\linewidth][c]{%
        \includegraphics[width=1.5\linewidth,keepaspectratio,trim={7cm 0.6cm 7cm 0.6cm},clip]{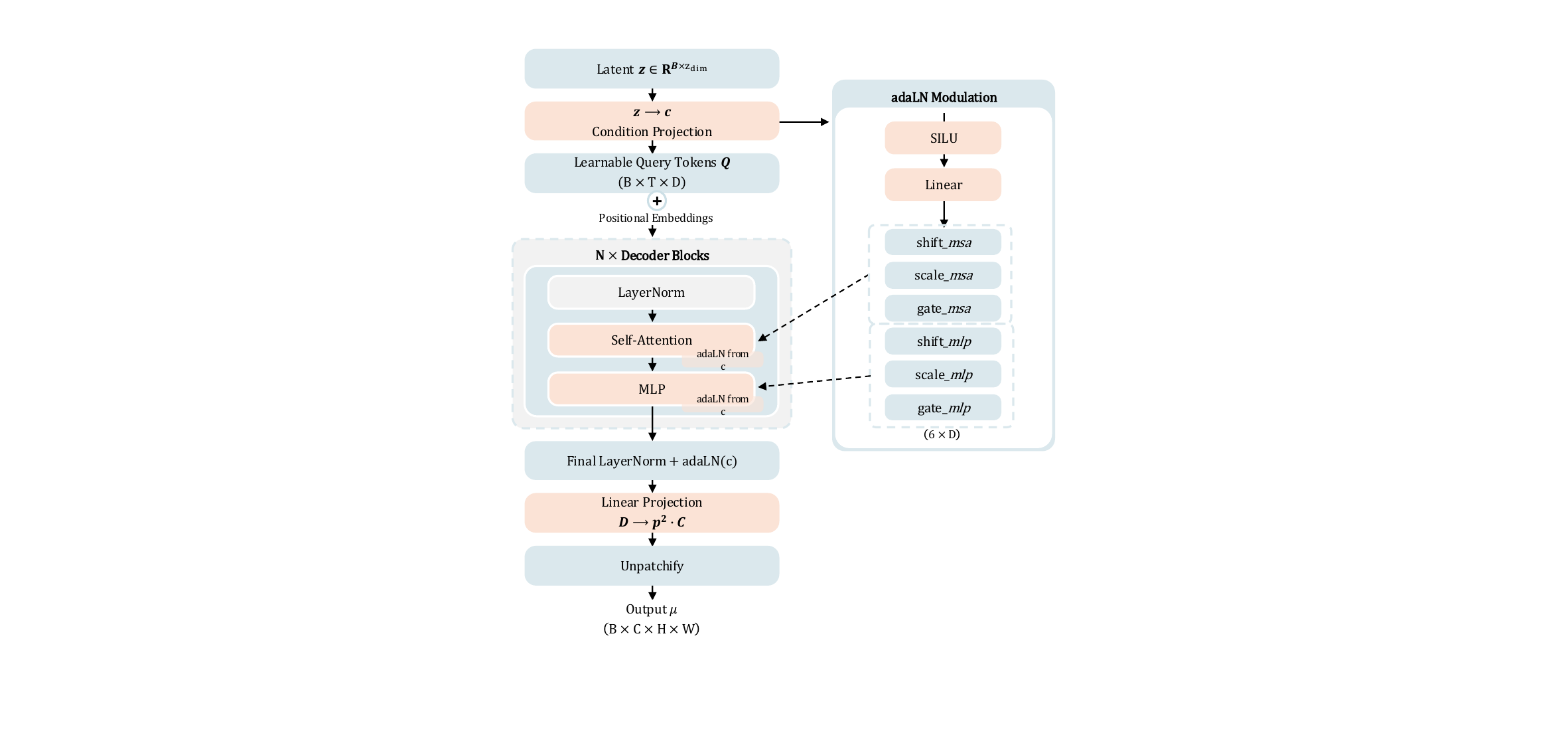}%
    }
    \caption{\textbf{SiT-like decoder architecture used in the ImageNet-128 SCFM setup.}}
    \label{fig:sit_decoder}
\end{figure}

Table~\ref{tab:imagenet_sit_xl2_enc_dec_setup} summarizes the ImageNet-128 latent SCFM configuration. We build on a pretrained Stable-Diffusion VAE \cite{rombach2022high}, using \texttt{stabilityai/sd-vae-ft-ema} to map $128{\times}128$ images into a $4$-channel latent image space. The SCFM encoder operates on these VAE latents using a SiT-XL/2 transformer backbone, while the decoder also uses a transformer architecture to map the learned $64$-dimensional structured latent representation back to the VAE latent image space. The decoder architecture is depicted in Figure~\ref{fig:sit_decoder}. The prior over $\mathbf{z}$ is modeled as a learnable GMM with $100$ components. The model is trained with a $\beta$-VAE endpoint objective and an $\ell_1$ reconstruction loss, and samples are generated with the Dormand--Prince ODE solver.

\begin{table}[!htbp]
    \centering
    \caption{\textbf{ImageNet-128 SiT-XL/2 latent model setup.}
        The encoder parameterizes the latent recognition pathway, while the decoder maps latent representations back to the VAE latent image space. Shared optimization and sampling settings are merged across both columns.}
    \label{tab:imagenet_sit_xl2_enc_dec_setup}
    \small
    \setlength{\tabcolsep}{4pt}
    \renewcommand{\arraystretch}{1.06}
    \begin{tabular}{lcc}
        \toprule
        \textbf{Hyperparameter} & \textbf{Encoder / recognition net}                                               & \textbf{Decoder} \\
        \midrule
        Role
                                & Latent encoder
                                & Latent decoder                                                                                      \\
        Backbone
                                & SiT-XL/2
                                & SiT-like decoder                                                                                    \\
        Input
                                & VAE latent image
                                & $\mathbf{z}\in\mathbb{R}^{64}$                                                                      \\
        Patch size
                                & $2{\times}2$
                                & --                                                                                                  \\
        In/out channels
                                & $4$ input channels
                                & $4$ output channels                                                                                 \\
        Hidden size
                                & $1152$
                                & $1152$                                                                                              \\
        Depth
                                & $28$
                                & $14$                                                                                                \\
        Attention heads
                                & $16$
                                & $16$                                                                                                \\
        MLP ratio
                                & $4.0$
                                & $4.0$                                                                                               \\
        Dropout
                                & Class dropout $0.1$
                                & Dropout $0.0$                                                                                       \\
        Conditioning
                                & \makecell[c]{AdaLN-Zero                                                                             \\time and null-token conditioning}
                                & AdaLN on $\mathbf{z}$                                                                               \\
        Latent dimension
                                & \multicolumn{2}{c}{$64$}                                                                            \\
        VAE
                                & \multicolumn{2}{c}{\texttt{stabilityai/sd-vae-ft-ema}, scaling factor $0.18215$}                    \\
        Prior
                                & \multicolumn{2}{c}{Learnable GMM prior with $100$ components}                                       \\
        Training steps
                                & \multicolumn{2}{c}{$400$K}                                                                          \\
        Batch size
                                & \multicolumn{2}{c}{$256$}                                                                           \\
        Optimizer
                                & \multicolumn{2}{c}{AdamW, $\beta_1=0.9$, $\beta_2=0.95$}                                            \\
        Learning rate
                                & \multicolumn{2}{c}{$1{\times}10^{-4}$ with linear decay}                                            \\
        Reconstruction loss
                                & \multicolumn{2}{c}{$\ell_1$ reconstruction loss}                                                    \\
        Endpoint objective
                                & \multicolumn{2}{c}{$\beta$-VAE with $\beta=2.8$}                                                    \\
        Gradient clipping
                                & \multicolumn{2}{c}{Maximum gradient norm $5.0$}                                                     \\
        EMA
                                & \multicolumn{2}{c}{Enabled, decay $0.9999$}                                                         \\
        Sampler
                                & \multicolumn{2}{c}{Dormand--Prince method of order 5 (Dopri5)}                                      \\
        \bottomrule
    \end{tabular}
\end{table}

\clearpage

\subsection{CIFAR-10 and ImageNet-128 Probing Setup}
\label{app:cifar_imagenet_probe_setup}

Table~\ref{tab:probe_settings_cifar_imagenet} summarizes the probing protocols for CIFAR-10 and ImageNet-128. For CIFAR-10, we follow the latent probing setup of \citet{zhang2022improvingvaebasedrepresentationlearning}. For ImageNet-128, we evaluate frozen latent representations with linear and nonlinear probes and compare SCFM against a frozen SD-VAE-FT-EMA encoder from Stable Diffusion~\citep{rombach2022high}. In both cases, probes are trained without data augmentation so that performance reflects information available in the frozen representation.

\begin{table}[!htbp]
    \centering
    \caption{\textbf{Probe settings for CIFAR-10 and ImageNet-128.}
        After training, frozen latent representations are evaluated with linear and nonlinear probes.}
    \label{tab:probe_settings_cifar_imagenet}
    \small
    \setlength{\tabcolsep}{4pt}
    \renewcommand{\arraystretch}{1.08}
    \begin{tabular}{lll}
        \toprule
        \textbf{Component} & \textbf{CIFAR-10}                     & \textbf{ImageNet-128} \\
        \midrule
        Representation
                           & $\mu_\phi(\mathbf{x})$
                           & $\mu_\phi(\mathbf{x})$                                        \\
        Target
                           & CIFAR-10 label
                           & ImageNet label                                                \\
        Data augmentation
                           & None
                           & None                                                          \\
        Model-selection split
                           & Not specified
                           & \makecell[l]{$5\%$ stratified holdout                         \\from ImageNet train} \\
        Evaluation split
                           & Standard test split
                           & ImageNet validation split                                     \\
        Linear probe
                           & Linear SVM
                           & \makecell[l]{Single linear layer                              \\on standardized features} \\
        Nonlinear probe
                           & \makecell[l]{2-layer MLP                                      \\hidden size 200, ReLU}
                           & \makecell[l]{Residual MLP                                     \\hidden $(4096,2048,1024)$\\GELU activations} \\
        Regularization
                           & \makecell[l]{BatchNorm                                        \\dropout $0.1$}
                           & \makecell[l]{LayerNorm                                        \\dropout $0.12$--$0.15$\\weight decay $10^{-4}$--$2{\times}10^{-4}$\\label smoothing $0.05$--$0.1$} \\
        Reporting
                           & Accuracy
                           & Top-1 / Top-5 accuracy                                        \\
        \bottomrule
    \end{tabular}
\end{table}

\section{Evaluation Metrics}
\label{app:metrics}
\subsection{Clustering Metrics}

We evaluate clustering performance using two standard metrics: clustering accuracy (ACC) and normalized mutual information (NMI).

\paragraph{Clustering accuracy (ACC).}
ACC measures the alignment between predicted cluster assignments and ground-truth class labels, accounting for the permutation ambiguity of cluster indices. Since cluster IDs are arbitrary, we first compute an optimal mapping $\pi$ from cluster IDs to class labels, and then measure the fraction of correctly assigned samples:
\begin{equation}
    \mathrm{ACC} = \frac{1}{N} \sum_{i=1}^{N} \mathbf{1}\{y_i = \pi(c_i)\},
\end{equation}
where $y_i$ denotes the ground-truth label and $c_i$ the predicted cluster assignment. When the number of clusters matches the number of classes, $\pi$ is obtained via Hungarian matching. For MFCVAE, we report facet-wise ACC using the cluster-to-label assignment induced by the dominant class within each cluster.

\paragraph{Normalized mutual information (NMI).}
NMI measures the statistical dependence between predicted clusters and ground-truth labels, and is invariant to label permutations. It is defined based on the mutual information $I(Y;C)$ between labels $Y$ and clusters $C$, normalized by their entropies. Using the arithmetic normalization adopted in our implementation:
\begin{equation}
    \mathrm{NMI}(Y, C) = \frac{2\, I(Y;C)}{H(Y) + H(C)}.
\end{equation}
NMI ranges from 0 to 1, where 0 indicates no mutual information and 1 corresponds to perfect agreement between clusters and labels.

\subsection{Disentanglement Metrics}

We evaluate disentanglement using the FactorVAE score \citep{kim2018disentangling} and the disentanglement component of the DCI framework \citep{eastwood2018framework}.

\paragraph{FactorVAE score.}
Let $r(x)\in\mathbb{R}^d$ denote the learned representation and let the data be generated by $K$ independent factors.
For each training example of the metric, we fix one ground-truth factor $k$, sample a batch $B_k$ while varying all other factors, encode the batch, and normalize each latent coordinate by its empirical standard deviation over the dataset:
\[
    s_j^2 = \mathrm{Var}_{p_{\mathrm{data}}}\!\left[r_j(x)\right].
\]
We then compute the within-batch variance of each normalized latent coordinate,
\[
    \hat v_j(B_k)
    =
    \mathrm{Var}_{x\in B_k}
    \left[
        \frac{r_j(x)}{s_j}
        \right],
\]
and predict the fixed factor using the index of the least-varying coordinate,
\[
    j^\star(B_k)=\arg\min_j \hat v_j(B_k).
\]
Accumulating many such pairs $(j^\star(B_k),k)$ yields a majority-vote classifier from latent dimensions to factors. The FactorVAE score is the resulting classification accuracy on held-out batches. Higher is better.

\paragraph{DCI disentanglement.}
The DCI framework measures disentanglement, completeness, and informativeness from an importance matrix relating latent coordinates to ground-truth factors.
Since Table~\ref{tab:disentangled_fm} reports only the disentanglement component, we denote that score by DCI.
Let $R\in\mathbb{R}_{\ge 0}^{L\times K}$ be a relative-importance matrix, where $R_{ij}$ measures how much latent coordinate $c_i$ contributes to predicting factor $z_j$, and each column is normalized so that $\sum_{i=1}^L R_{ij}=1$.
For each latent coordinate, define the normalized row distribution
\[
    P_{ij}
    =
    \frac{R_{ij}}{\sum_{k=1}^K R_{ik}},
    \qquad
    D_i
    =
    1 - H_K(P_{i\cdot}),
\]
where
\[
    H_K(P_{i\cdot})
    =
    -\sum_{j=1}^K P_{ij}\log_K P_{ij}.
\]
The overall disentanglement score is the importance-weighted average
\[
    \mathrm{DCI}
    =
    \sum_{i=1}^L \rho_i D_i,
    \qquad
    \rho_i
    =
    \frac{1}{K}\sum_{j=1}^K R_{ij}.
\]
This score is highest when each latent coordinate is primarily predictive of a single generative factor.

\subsection{Generative Quality Metric}

\paragraph{Fr\'echet inception distance (FID).}
To evaluate image quality and diversity, we use FID \citep{heusel2017gans}.
Let $(m_r, C_r)$ and $(m_g, C_g)$ denote the empirical mean and covariance of Inception features extracted from real and generated samples, respectively.
FID is defined as
\[
    \mathrm{FID}
    =
    \|m_r - m_g\|_2^2
    +
    \operatorname{Tr}
    \left(
    C_r + C_g - 2(C_r C_g)^{1/2}
    \right).
\]
Lower FID indicates that generated samples are closer to real data in the feature space, reflecting both fidelity and diversity.

\subsection{Downstream Representation Metrics}

\paragraph{Linear and nonlinear probing.}
To evaluate whether the learned latent representation retains class-relevant semantic information, we freeze the encoder and train supervised probes on top of the latent code. Let
\[
    \mathbf{z}_i = r_\phi(\mathbf{x}_i)
\]
denote the frozen representation of image \(\mathbf{x}_i\). A linear probe trains a single linear classifier
\[
    h(\mathbf{z}) = W\mathbf{z} + b
\]
using labeled training examples, while keeping \(r_\phi\) fixed. A nonlinear probe uses a small multilayer classifier on top of the same frozen representation.

Probe performance is measured by Top-\(k\) classification accuracy on the held-out test set:
\[
    \mathrm{Top}\text{-}k
    =
    \frac{1}{N}
    \sum_{i=1}^{N}
    \mathbf{1}
    \left\{
    y_i \in \operatorname{TopK}\bigl(h(\mathbf{z}_i), k\bigr)
    \right\},
\]
where \(\operatorname{TopK}(h(\mathbf{z}_i), k)\) denotes the set of \(k\) classes with the highest predicted scores. The usual classification accuracy corresponds to the special case \(k=1\):
\[
    \mathrm{Acc}
    =
    \mathrm{Top}\text{-}1
    =
    \frac{1}{N}
    \sum_{i=1}^{N}
    \mathbf{1}
    \left\{
    \arg\max_c h_c(\mathbf{z}_i) = y_i
    \right\}.
\]
For CIFAR-10, we report Top-1 accuracy. For ImageNet-128, we report both Top-1 and Top-5 accuracy
\section{Additional Results and Diagnostics}
\label{app:additional-results}
This section collects supplementary diagnostics for the representation-learning experiments in Section~\ref{sec:exp-structured-latents} and the image-generation experiments in Section~\ref{sec:exp-image-generation}.

\subsection{MNIST Clustering and Latent Diagnostics}
\label{app:mnist_latent_diagnostics}

\begin{figure}[!htbp]
    \centering
    \includegraphics[width=\linewidth]{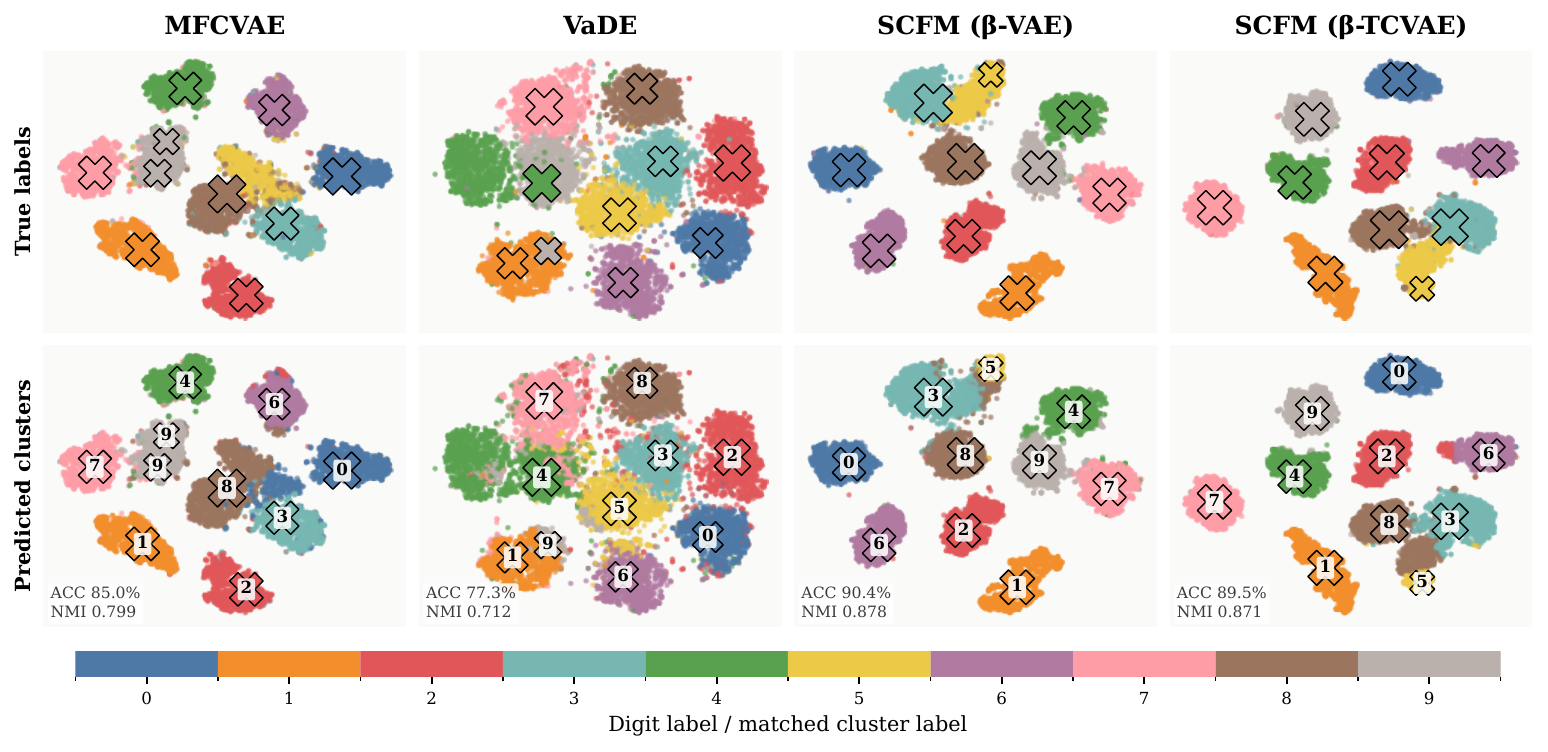}
    \caption{t-SNE visualization of the learned latent spaces on MNIST.
        Each column corresponds to a different model.
        The top row colors points by digit label, and the bottom row colors the same embeddings by matched cluster assignment.}
    \label{fig:mnist_latent_comparison}
\end{figure}

\paragraph{Latent clustering.}
Figure~\ref{fig:mnist_latent_comparison} compares the latent spaces learned by MFCVAE, VaDE, and the two SCFM variants. Columns correspond to different models. The top row colors points by ground-truth digit label, while the bottom row colors the same embeddings by predicted cluster assignment after cluster-to-label matching. The SCFM variants produce more compact and label-consistent clusters than the VAE-based baselines, with SCFM ($\beta$-VAE) achieving the best clustering performance in Table~\ref{tab:mnist_results}.

\begin{table}[!htbp]
    \caption{\textbf{Clustering performance on MNIST} (mean $\pm$ std). Baseline methods are VaDE~\citep{jiang2017variational} and MFCVAE~\citep{falck2021multi}.}
    \label{tab:mnist_results}
    \centering
    \small
    \begin{tabular}{lcc}
        \toprule
        \textbf{Model}       & \textbf{NMI} $\uparrow$   & \textbf{ACC} $\uparrow$   \\
        \midrule
        VaDE                 & 79.93 $\pm$ 0.77          & 77.33 $\pm$ 0.33          \\
        MFCVAE               & 71.22 $\pm$ 0.68          & 85.43 $\pm$ 0.54          \\
        SCFM ($\beta$-VAE)   & \textbf{87.84 $\pm$ 0.86} & \textbf{90.44 $\pm$ 0.60} \\
        SCFM ($\beta$-TCVAE) & 87.11 $\pm$ 0.47          & 89.52 $\pm$ 0.84          \\
        \bottomrule
    \end{tabular}
\end{table}
\clearpage

\subsection{CIFAR-10 Component-Level Generation Diagnostics}
\label{app:cifar10_generation}

\begin{figure*}[!htbp]
    \centering
    \includegraphics[width=\textwidth]{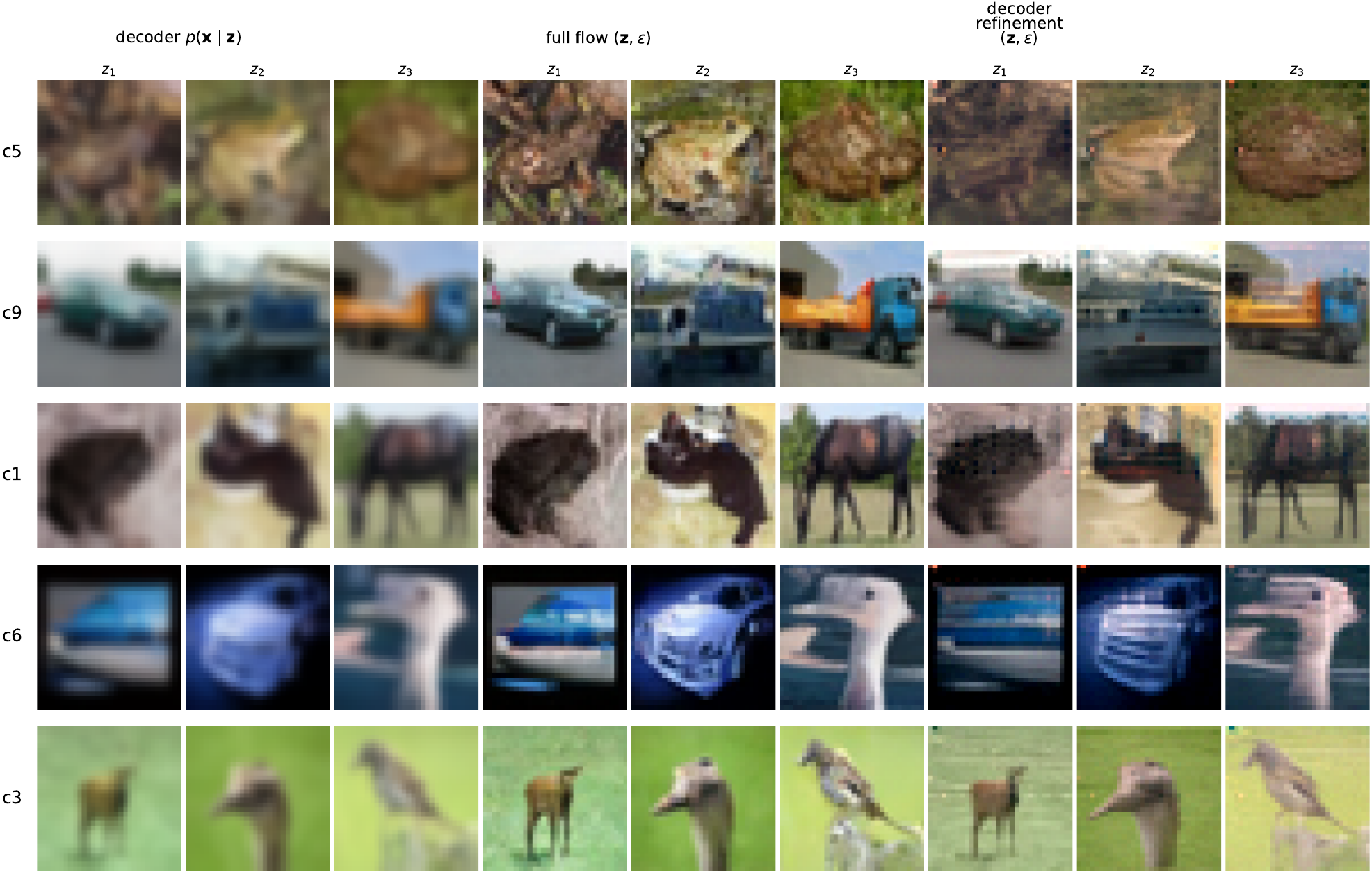}
    \caption{\textbf{Representative CIFAR-10 mixture components learned by SCFM.}
        Each row corresponds to one learned GMM component. The first three columns show decoder-only samples $p_\theta(\mathbf{x}\mid\mathbf{z})$ for latent draws $\mathbf{z}$ assigned to the component. The next three columns show full-flow samples obtained from the structured source $(\mathbf{z},\varepsilon)$. The last three columns show decoder-initialized refinement samples initialized from the decoder output and refined by integrating from $t_0=0.8$ with $5$ NFE. Although the components are not label-pure, they capture coherent appearance modes such as shared color palettes, backgrounds, object scale, and coarse geometry, while flow refinement improves visual detail without destroying component coherence.}
    \label{fig:cifar_component_rows}
\end{figure*}

Figure~\ref{fig:cifar_component_rows} analyzes how the learned GMM prior organizes CIFAR-10 samples and how decoder-initialized refinement affects generation quality. Each row corresponds to one learned component. For each component, we compare decoder-only samples from the learned latent prior, full-flow samples obtained by integrating from the structured source, and decoder-initialized refinement samples initialized from the decoder and refined by the flow. The components are not class-pure, which is expected for CIFAR-10 without label supervision. Instead, samples within a component often share coherent appearance statistics, such as color palette, background structure, object scale, or coarse geometry. This suggests that the learned prior captures appearance modes rather than directly recovering semantic classes.

The comparison across sampling modes highlights the cooperative role of the decoder and the flow. Decoder-only samples already capture coarse component-specific structure, while full-flow sampling improves realism and fine detail. Decoder-initialized refinement preserves the coarse structure proposed by the decoder while adding sharper edges and more realistic textures. In the last three columns, refinement starts from the decoder output at $t_0=0.8$ and uses only $5$ NFE, compared with $50$ NFE for full-flow sampling, yet remains visually close to the full-flow samples. This supports decoder-initialized refinement as a lower-cost sampling mode.

\vspace{5em}

\subsection{CIFAR-10 Latent Probe Diagnostics}
\label{app:cifar10_latent_probe_diagnostics}

For CIFAR-10, we evaluate how much class-relevant information is accessible from the learned latent code. We train linear and nonlinear probes on frozen representations, without updating the SCFM encoder or using data augmentation.

\begin{figure*}[!htbp]
    \centering
    \includegraphics[width=\textwidth]{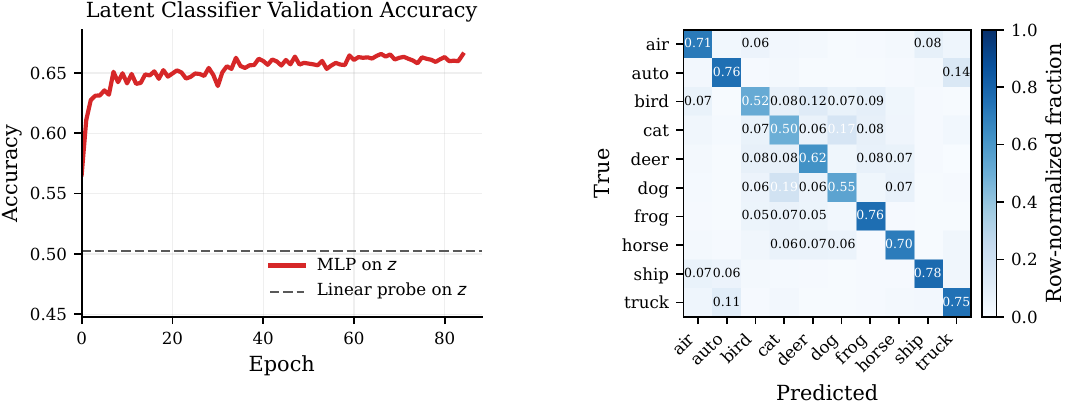}
    \caption{\textbf{CIFAR-10 latent classifier diagnostics.}
        Left: validation accuracy of an MLP trained on the frozen latent code $\mathbf{z}$, with the linear-probe accuracy shown as a dashed baseline.
        Right: row-normalized confusion matrix for the latent MLP classifier.
        The representation separates vehicle classes reliably, while most residual confusion occurs among animal categories with similar pose and texture statistics.}
    \label{fig:cifar_latent_classifier}
\end{figure*}

For SCFM, probe results are averaged over five independently trained models with different random seeds. Non-SCFM baseline results are reported from \citet{zhang2022improvingvaebasedrepresentationlearning}. Table~\ref{tab:cifar_compare} reports CIFAR-10 probe accuracies, while Figure~\ref{fig:cifar_latent_classifier} shows latent-classifier diagnostics for a representative nonlinear probe.

\begin{table}[!htbp]
    \caption{\textbf{CIFAR-10 probe accuracy from learned latent representations}. Linear and nonlinear probe results are shown. SCFM is reported as the mean over five seeds, while the non-SCFM baseline numbers are taken from \citet{zhang2022improvingvaebasedrepresentationlearning}.}
    \label{tab:cifar_compare}
    \centering
    \small
    \begin{tabular}{lcc}
        \toprule
        \textbf{Model} & \textbf{Linear Probe} $\uparrow$ & \textbf{Nonlinear Probe} $\uparrow$ \\
        \midrule
        VAE            & 39.59                            & 54.61                               \\
        AAE            & 37.76                            & 52.81                               \\
        BiGAN          & 44.90                            & 52.54                               \\
        DIM            & 49.62                            & \textbf{69.13}                      \\
        SCFM (ours)    & \textbf{50.26}                   & 66.58                               \\
        \bottomrule
    \end{tabular}
\end{table}

\subsection{ImageNet-128 Latent Probe Diagnostics}

We evaluate whether the representation gains observed on smaller datasets persist at ImageNet scale. For ImageNet-128, we freeze the latent representation and train linear and nonlinear probes without data augmentation. Probe performance is measured using Top-1 and Top-5 accuracy, which quantify how much ImageNet class information is accessible from the latent space. We compare SCFM against a frozen Stable-Diffusion VAE encoder baseline using SD-VAE-FT-EMA~\citep{rombach2022high}.
Table~\ref{tab:imagenet_latent_probe_scfm} shows that SCFM improves over the SD-VAE baseline for both probe families, with the largest gains under the nonlinear probe. This suggests that the SCFM latent space retains substantially more class-relevant information than the frozen VAE encoder alone, while still being learned through the generative training objective.

\begin{table}[!htbp]
    \centering
    \small
    \caption{\textbf{ImageNet-128 latent-space probing.}
        Frozen latent representations are evaluated with linear and nonlinear probes.
        The VAE baseline uses the frozen SD-VAE-FT-EMA encoder from Stable Diffusion~\citep{rombach2022high}.}
    \label{tab:imagenet_latent_probe_scfm}
    \begin{tabular}{llcc}
        \toprule
        \textbf{Method} & \textbf{Probe} & \textbf{Top-1 Acc.} $\uparrow$ & \textbf{Top-5 Acc.} $\uparrow$ \\
        \midrule
        SD-VAE-FT-EMA   & Linear         & 8.00                           & 13.23                          \\
        SD-VAE-FT-EMA   & Non-linear     & 12.34                          & 26.54                          \\
        \midrule
        SCFM (ours)     & Linear         & \textbf{9.07}                  & \textbf{23.14}                 \\
        SCFM (ours)     & Non-linear     & \textbf{27.96}                 & \textbf{53.08}                 \\
        \bottomrule
    \end{tabular}
\end{table}

\vspace{8em}

\section{Additional Qualitative Samples}
\label{app:additional_qualitative}

\begin{figure}[!htbp]
    \centering
    \begin{subfigure}{0.95\linewidth}
        \centering
        \includegraphics[width=\linewidth]{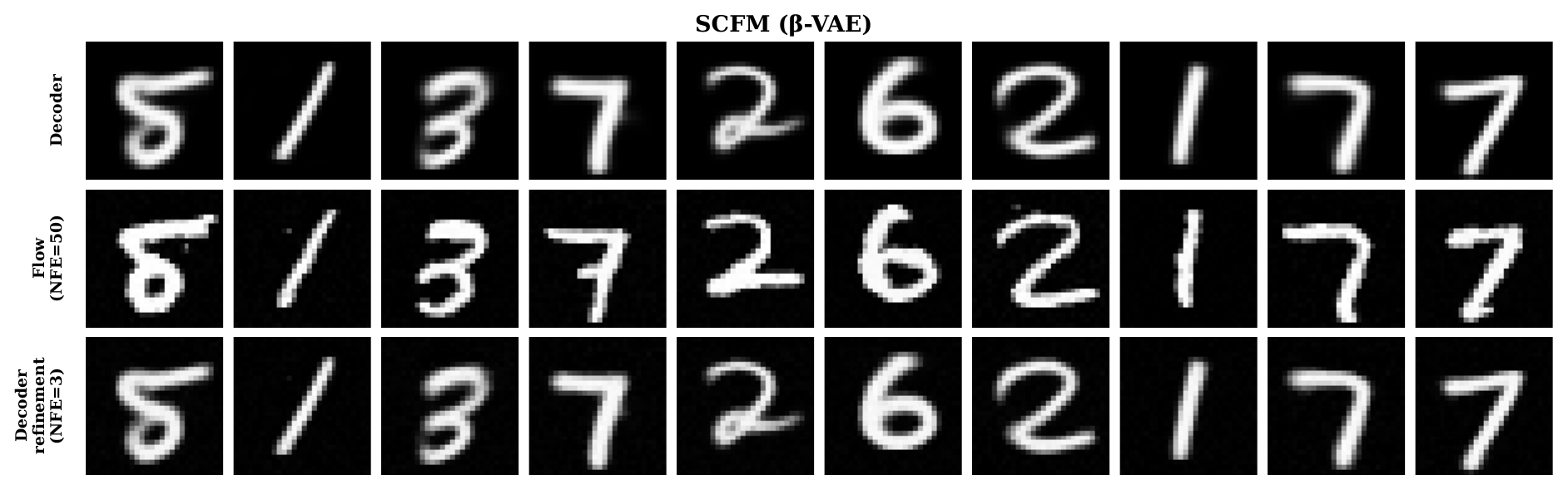}
    \end{subfigure}

    \vspace{0.5em}

    \begin{subfigure}{0.95\linewidth}
        \centering
        \includegraphics[width=\linewidth]{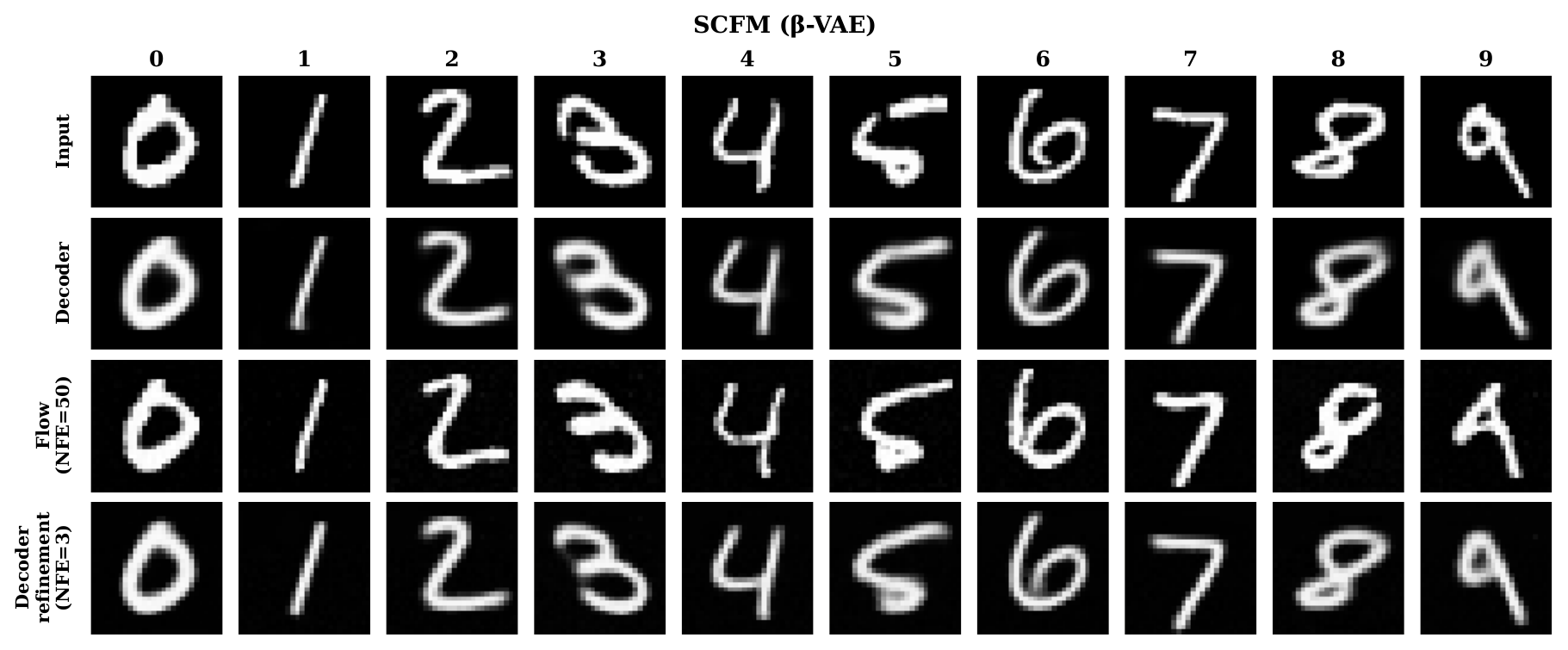}
    \end{subfigure}

    \caption{\textbf{MNIST generation and reconstruction with SCFM ($\beta$-VAE).}
        Top panel: generation with decoder-only samples from the learned prior, full-flow samples obtained by integrating from the source prior with 50 function evaluations (NFE), and decoder-initialized refinement initialized from the decoder output at $t_0=0.9$ and refined with 3 NFE.
        Bottom panel: reconstruction with input images, decoder-only reconstructions, full-flow reconstructions obtained by integrating from the encoder-induced source with 50 NFE, and decoder-initialized refinement initialized from the decoder output at $t_0=0.9$ and refined with 3 NFE.}
    \label{fig:mnist_beta_vae_modes}
\end{figure}

\paragraph{MNIST Generation and reconstruction modes.}
Figure~\ref{fig:mnist_beta_vae_modes} evaluates whether the learned latent space remains useful for generation and reconstruction. Decoder-only samples provide an endpoint proposal, full-flow samples integrate from the learned source prior, and decoder-initialized refinement starts from the decoder output at $t_0=0.9$ and applies only $3$ flow evaluations. The results show that the decoder captures the digit identity, while short flow refinement improves visual consistency. This illustrates the cooperative role of the VAE endpoint branch and the flow-matching transport in SCFM.

\begin{figure}[!htbp]
    \centering
    \includegraphics[width=\linewidth]{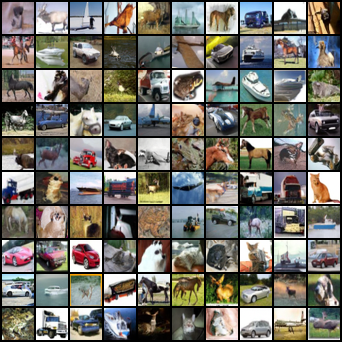}
    \caption{\textbf{Additional uncurated CIFAR-10 samples from SCFM.}
        Samples are generated with full-flow sampling from the learned structured source prior.}
    \label{fig:cifar_fake_samples}
\end{figure}

\begin{figure}[!htbp]
    \centering
    \includegraphics[width=\linewidth]{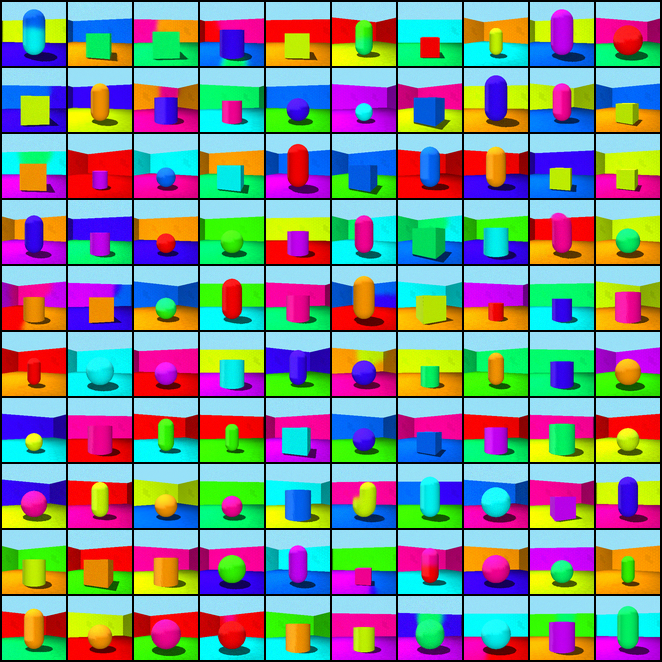}
    \caption{\textbf{Additional uncurated Shapes3D samples from SCFM.}
        Samples are generated with full-flow sampling from the learned structured source prior.}
    \label{fig:shapes3d_fake_samples}
\end{figure}

\begin{figure}[!htbp]
    \centering
    \includegraphics[width=\linewidth]{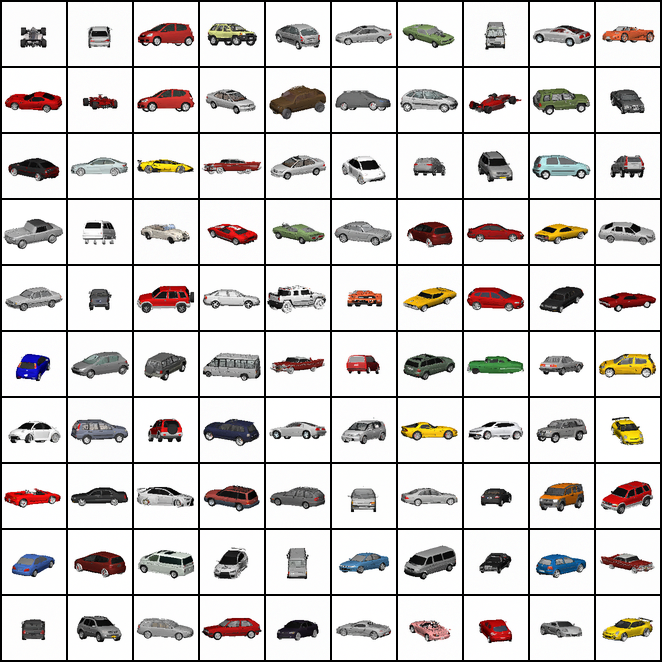}
    \caption{\textbf{Additional uncurated Cars3D samples from SCFM.}
        Samples are generated with full-flow sampling from the learned structured source prior.}
    \label{fig:cars3d_fake_samples}
\end{figure}

\begin{figure}[!htbp]
    \centering
    \includegraphics[width=\linewidth]{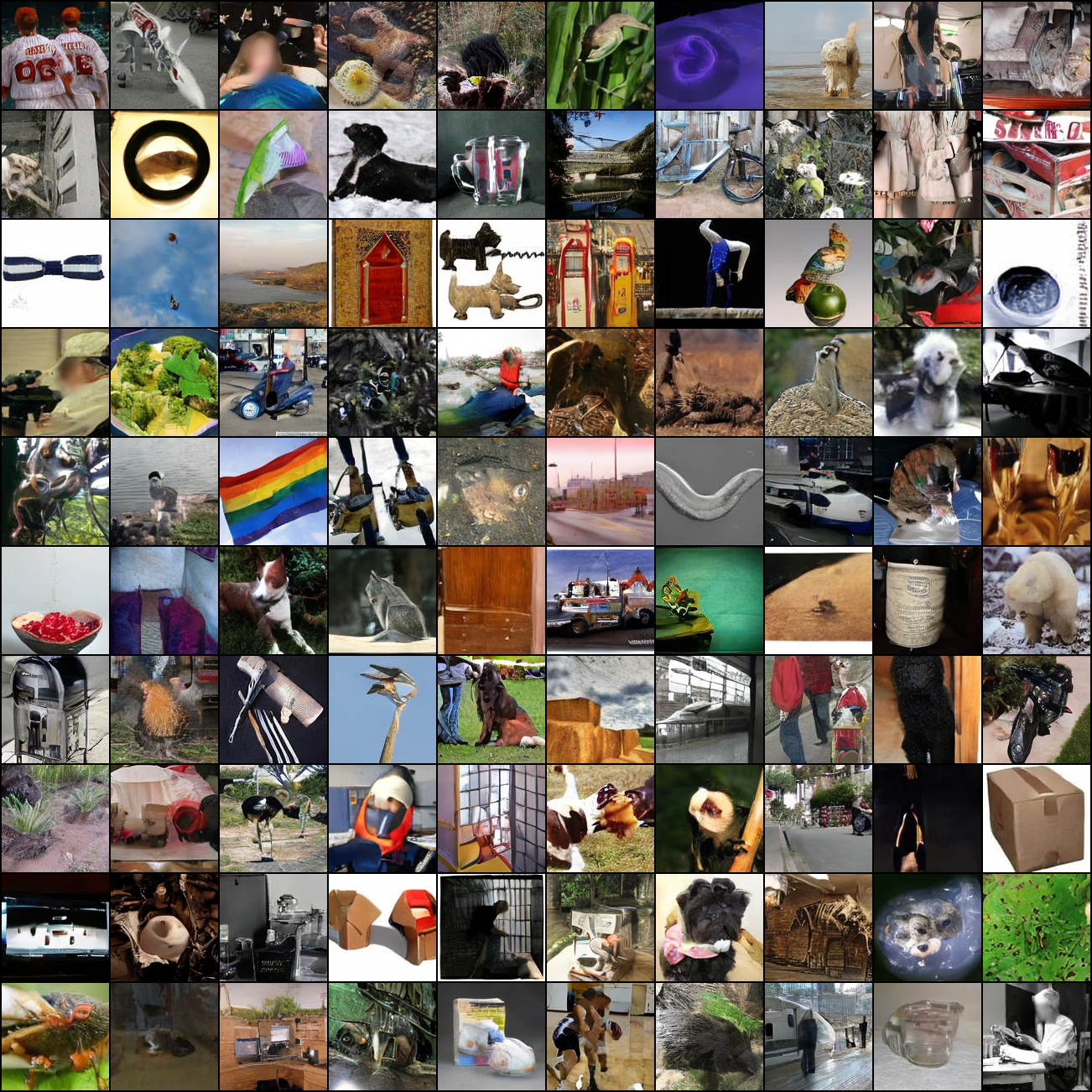}
    \caption{\textbf{Additional uncurated ImageNet-128 samples from SCFM.}
        Samples are generated with full-flow sampling from the learned structured source prior.}
    \label{fig:imagenet128_fake_samples}
\end{figure}

The rest of this section provides additional qualitative samples for the SCFM models.
Figures~\ref{fig:cifar_fake_samples}, \ref{fig:shapes3d_fake_samples}, \ref{fig:cars3d_fake_samples}, and \ref{fig:imagenet128_fake_samples} show additional uncurated full-flow samples for CIFAR-10, Shapes3D, Cars3D, and ImageNet-128, respectively.

\ifpreprintmode\else
  \clearpage
\section{Broader Impacts}
\label{app:broader-impact}

This work studies how to equip flow-matching generators with structured latent variables that improve unsupervised representation learning and provide more interpretable control over generation. Potential positive impacts include generative models whose latent structure is easier to analyze, representations that are more useful for downstream tasks without requiring labels, and more transparent generative systems in domains where understanding semantic structure matters.

At the same time, SCFM remains a general-purpose image generator and inherits the risks associated with modern generative models. In particular, improved latent structure and controllable semantic variation could be misused to manipulate generated content more deliberately, while the unsupervised latent variables may absorb and reproduce biases present in the training data. Because the method is trained on large-scale image datasets, privacy-sensitive or socially biased correlations in the data may also be reflected in the learned representation. We therefore view SCFM as a research contribution rather than a deployment-ready system and emphasize that any real-world use should account for dataset bias, privacy, and misuse risks.

\fi

\ifpreprintmode\else
  \clearpage
  \newpage
  \input{checklist.tex}
\fi

\end{document}